\documentclass[runningheads]{llncs}

\usepackage[year=2026]{eccv}

\makeatletter
\renewcommand\paragraph{\@startsection{paragraph}{4}{\z@}%
  {0.8ex \@plus 0.2ex \@minus 0.2ex}%
  {-0.8em}%
  {\normalfont\normalsize\bfseries}}
\makeatother

\usepackage{eccvabbrv}

\usepackage{graphicx}
\usepackage{booktabs}
\usepackage[normalem]{ulem}

\usepackage{xcolor}         
\usepackage{colortbl}

\usepackage{multirow}
\usepackage{booktabs,tabularx,makecell,threeparttable,pifont,siunitx}
\usepackage{soul}
\newcommand{\cmark}{\ding{51}}   
\newcommand{\xmark}{\ding{55}}   
\definecolor{LightGray}{rgb}{0.9,0.9,0.9}
\usepackage{hyperref}

\usepackage{orcidlink}

\begin{document}

\title{SKEL-CF: Coarse-to-Fine Biomechanical Skeleton and Surface Mesh Recovery} 

\titlerunning{SKEL-CF}


\author{
Da Li$^{1,2*}$\orcidlink{0009-0007-5452-1089}, 
Jiping Jin$^{1,3*}$\orcidlink{0009-0003-0693-7216},
Xuanlong Yu$^{1}$\orcidlink{0000-0001-8438-5187}, 
Wei Liu$^{5}$\orcidlink{0009-0008-0452-1336}, 
Xiaodong Cun$^{4}$\orcidlink{0000-0003-3607-2236}, \\ 
Kai Chen$^{5}$\orcidlink{0009-0007-2686-9336}, 
Rui Fan$^{3}\orcidlink{0000-0002-6334-6282}$, 
Jiangang Kong$^{5}$\orcidlink{0009-0002-6018-8642}, 
Xi Shen$^{1,\dagger}$\orcidlink{0000-0001-8043-9117} \\[1mm]
\small $^{*}$Equal contribution \qquad $^{\dagger}$Corresponding author \\[1mm]
$^1$Intellindust AI Lab \qquad
$^2$Shenzhen University \qquad
$^3$ShanghaiTech University \\
$^4$GVC Lab, Great Bay University \qquad
$^5$Didi Chuxing Co.Ltd
}

\authorrunning{D. Li, J. Jin, X. Shen et al.}

\institute{}

\maketitle

\begin{abstract}
  
    Parametric 3D human models such as SMPL have driven significant advances in human pose and shape estimation, yet their simplified kinematics limit biomechanical realism. The recently proposed SKEL model addresses this limitation by re-rigging SMPL with an anatomically accurate skeleton. However, estimating SKEL parameters directly remains challenging due to limited training data, perspective ambiguities, and the inherent complexity of human articulation. In this work, we propose SKEL-CF, a new framework for estimating SKEL parameters. SKEL-CF adopts a standard transformer-based encoder–decoder architecture. The encoder first produces coarse predictions of the camera extrinsics and SKEL parameters. The decoder then iteratively refines these predictions across multiple layers, with explicit and auxiliary supervision applied at each layer. To provide anatomically consistent training data, we convert the existing SMPL-based dataset into a SKEL-aligned version, called HMR-SKEL. This new dataset offers high-quality supervision for SKEL estimation. In addition, to reduce depth and scale ambiguity, we explicitly incorporate camera intrinsic estimation into the SKEL-CF pipeline and show that it is important for accurate reconstruction. Extensive experiments validate the effectiveness of the proposed design. On the challenging MOYO dataset, SKEL-CF achieves 85.0 MPJPE / 51.4 PA-MPJPE, significantly outperforming the previous SKEL-based state-of-the-art HSMR (104.5 / 79.6). These results establish SKEL-CF as a promising framework for human motion analysis, facilitating the use of computer vision techniques in biomechanics-related analysis. Our implementation is available on the project page: \url{https://pokerman8.github.io/SKEL-CF/}.
\end{abstract}
\section{Introduction}

\begin{figure*}[t]
    \centering
    \includegraphics[width=\textwidth]{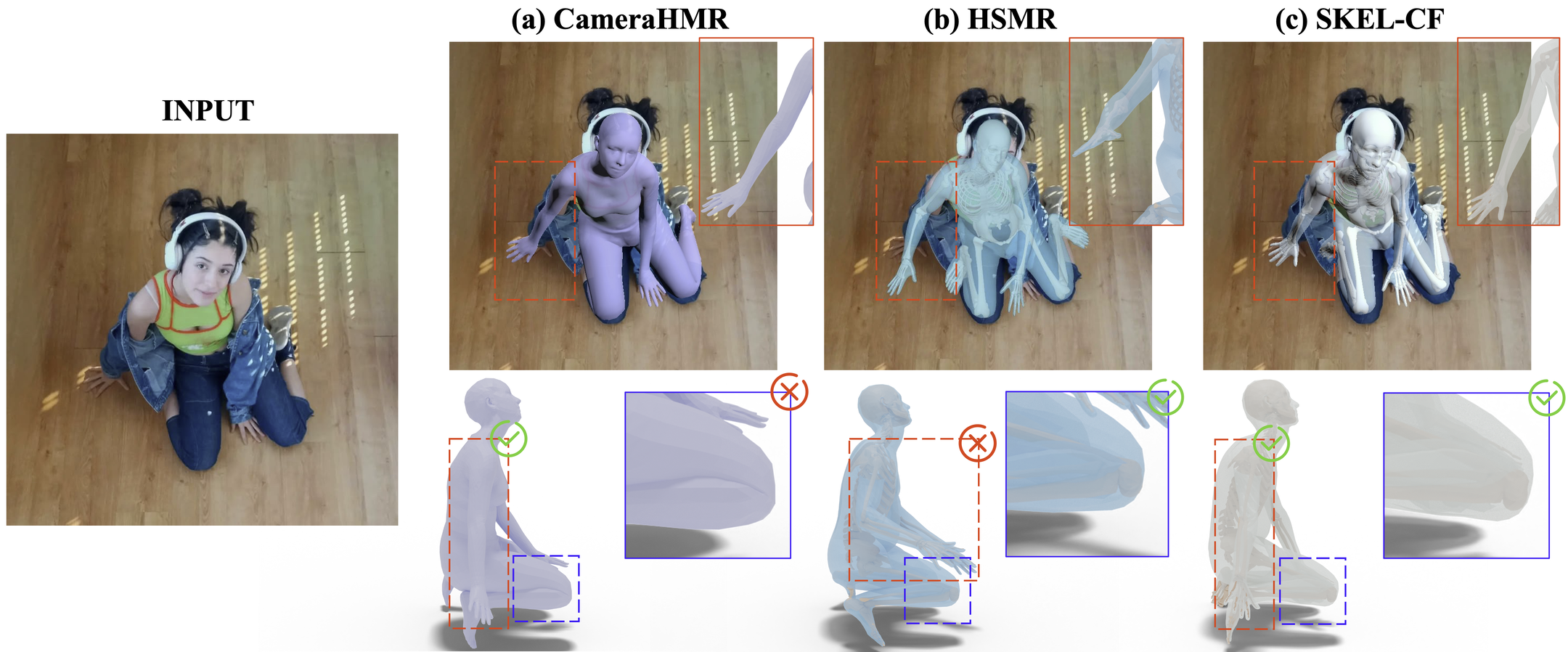}
    \caption{\textbf{Visualization of SKEL-CF on web images.} Compared with SMPL-based state-of-the-art CameraHMR~\cite{patel2024camerahmr}, our SKEL-based model produces more natural joint motions (see the side-view zoomed knee in \textcolor{blue}{blue}). Compared with HSMR~\cite{xia2025hsmr}, the state-of-the-art SKEL-based method, SKEL-CF achieves more accurate skeleton and mesh reconstruction (see zoomed hand in \textcolor{orange}{orange}).}
    \label{fig:teaser}
\end{figure*}

In recent years, 3D human pose and shape estimation has made remarkable progress, enabling a wide range of downstream applications~\cite{fu2024humanplus,he2024human2humannoid,li2024ohumanoidrobot,weng2022humannerf,tevet2022human}. However, adoption in biomechanics, a domain where such techniques could be particularly impactful, remains limited. This gap arises because existing parametric models, such as SMPL~\cite{loper2015smpl}, SMPL-X~\cite{smpl-x}, and GHUM~\cite{xu2020ghum}, fall short of the stringent biomechanical requirements. Their simplified kinematics and loosely constrained axis–angle representations often compromise biomechanical realism, especially in complex articulations such as deep squats or yoga poses~\cite{xia2025hsmr}. In these scenarios, the estimated parameters must correspond to anatomically accurate skeletons, respect joint limits, and ensure physically plausible motion with high precision. As illustrated in Fig.~\ref{fig:teaser} (a), even the state-of-the-art SMPL-based method CameraHMR~\cite{patel2024camerahmr} can produce anatomically implausible results, such as unnatural knee bending.

To address these limitations, the recently proposed SKEL~\cite{SKEL} model redefines the foundation of human representation by re-rigging SMPL with an anatomically accurate skeleton and realistic joint constraints. This makes it a promising step toward bridging computer vision and biomechanics, offering a representation that is both visually coherent and biomechanically faithful.
HSMR~\cite{xia2025hsmr} is the first transformer-based method to reconstruct a human body in the SKEL space from a single image. It converts the predicted SMPL ground truth from existing datasets into SKEL pseudo ground truth for supervised training. However, estimating SKEL from a single image remains challenging because it requires recovering a more constrained 3D structure than SMPL 
under the inherent depth ambiguity of monocular input. Consequently, the predictions often lack accuracy  (see Fig.~\ref{fig:teaser} (b)), and empirical metrics such as MPJPE remain inferior to SMPL-based models.


In this work, we advance SKEL-based human mesh recovery by proposing SKEL-CF, a new framework for estimating SKEL parameters. Similar to HSMR~\cite{xia2025hsmr}, SKEL-CF follows a standard transformer-based encoder–decoder architecture. However, unlike HSMR, which applies supervision only at the final layer, we adopt a coarse-to-fine estimation strategy, which is inspired by DETR~\cite{detr,zhu2020deformable}: the encoder predicts initial coarse camera extrinsics and SKEL parameters. The decoder then predicts the residual to refine the prediction. Moreover, by iteratively refining the predictions layer by layer in the decoder, with supervision applied at each decoder layer, we empirically observe further improvement. To handle diverse camera viewpoints, we incorporate camera intrinsic estimation following CameraHMR~\cite{patel2024camerahmr}. This improves robustness across different perspectives and reduces depth and scale ambiguity. Finally, leveraging the high-quality SMPL estimations from CameraHMR~\cite{patel2024camerahmr}, we generate large-scale and high-fidelity SKEL annotations to construct HMR-SKEL dataset. This dataset provides reliable supervision and enables more effective training for SKEL estimation. We position SKEL-CF as a SKEL-specific recovery pipeline that jointly addresses annotation quality, perspective ambiguity, and constrained biomechanical parameter regression. 

Empirically, SKEL-CF achieves substantial improvements across multiple benchmarks. On the challenging MOYO dataset~\cite{moyo}, it achieves 85.0 MPJPE and 51.4 PA-MPJPE, significantly outperforming the previous SKEL-based state-of-the-art HSMR~\cite{xia2025hsmr}, which reports 104.5 and 79.6, respectively. To further evaluate performance on challenging motions, we construct \textit{MOYO-HARD} by removing the first and last 25\% of frames from each sequence, where subjects are typically in static poses. On this subset, SKEL-CF achieves 90.0 MPJPE and 61.5 PA-MPJPE, compared with 120.0 and 97.7 for HSMR. In addition, SKEL-CF achieves comparable or better performance than leading SMPL-based methods such as CameraHMR~\cite{patel2024camerahmr} on 3DPW~\cite{3dpw}, EMDB~\cite{kaufmann2023emdb}, and MOYO~\cite{moyo}, while producing more biomechanically realistic and visually consistent human meshes (Fig~\ref{fig:teaser}(c)). We also perform extensive ablation studies to validate the effectiveness of each component in SKEL-CF.

To conclude, our key contributions are summarized as follows:

\begin{itemize}
\item We present SKEL-CF as an integrated SKEL-based recovery framework that combines high-quality SKEL supervision, perspective-aware camera conditioning, and coarse-to-fine SKEL regression for faithful mesh and skeleton recovery.
\item We introduce HMR-SKEL, a large-scale, high-quality dataset with anatomically consistent SKEL annotations converted from CameraHMR-refined SMPL annotations~\cite{patel2024camerahmr}. This dataset provides reliable supervision and unified skeleton--mesh alignment, enabling stronger SKEL-based model training.
\item Our ablation studies demonstrate that camera intrinsic estimation reduces perspective and scale ambiguity, coarse-to-fine estimation (encoder prediction followed by decoder refinement) improves performance, and iterative refinement with layer-wise decoder supervision further enhances robustness, particularly on challenging articulated motions such as MOYO-HARD. 

\end{itemize}

Our implementation is available on the project page: \url{https://pokerman8.github.io/SKEL-CF/}.

\section{Related Work}
\label{sec:Related_works}

\paragraph{Parametric human body model.} Parametric human body models are widely used for 3D human shape and pose estimation, as they offer a structured low-dimensional representation of human geometry. SMPL~\cite{loper2015smpl} represents the human mesh with fixed topology, using low-dimensional shape and pose parameters with linear blend skinning and pose-dependent corrective shapes for realistic articulation. SMPL-H~\cite{smpl-h} augments this framework with articulated hands, while SMPL-X~\cite{smpl-x} further integrates the FLAME~\cite{FLAME-HEAD} facial model, yielding a unified representation of body, hands, and expressive face. While the SMPL family provides a compact and effective representation of the human body surface, it employs a simplified kinematic structure that deviates from the anatomical skeletal system of real humans, allowing anatomically implausible poses (e.g., the knee can exhibit unrealistically free rotation, see Fig.~\ref{fig:teaser} (b)). SKEL~\cite{SKEL} introduces anatomically accurate joint definitions and a bone hierarchy embedded within a parametric human body model, overcoming limitations of simplified kinematic structures in models such as SMPL. By estimating parameters that are explicitly compatible with biomechanical skeletons, SKEL enforces realistic joint limits and produces physically plausible motion, leading to more accurate and reliable pose reconstruction. 

\paragraph{Human mesh recovery.} The lack of 3D annotations for in-the-wild images remains a major obstacle in Human mesh recovery. SMPLify~\cite{smplify} mitigates this by iteratively optimizing SMPL~\cite{loper2015smpl} parameters from 2D keypoints, yielding accurate results but at high computational cost. HMR~\cite{kanazawa2018hmr} introduces the first end-to-end framework to predict 3D pose and shape from in-the-wild images using only 2D supervision, where adversarial learning with mocap data provides strong pose and shape priors to compensate for the missing 3D information. 4DHuman~\cite{goel20234dhuman} constructs a large-scale dataset by fitting approximately 3 million images with ProHMR~\cite{prohmr}, generating pseudo-ground-truth (p-GT) 3D annotations alongside 2D keypoints, and further upgrades the network architecture from CNN to Vision Transformer (ViT). However, TokenHMR~\cite{dwivedi2024tokenhmr} identifies a mismatch between 3D and 2D keypoints, where enforcing 2D keypoint alignment can degrade 3D evaluation metrics, partly due to the use of fixed camera intrinsics and p-GT annotations in the 4DHuman~\cite{goel20234dhuman} dataset. To address this, TokenHMR~\cite{dwivedi2024tokenhmr} introduces the TALS loss, which down-weights the 2D keypoint loss when the L2 distance between predicted and ground-truth keypoints falls within a predefined threshold, thus preventing overfitting to noisy p-GT annotations. In addition, it leverages a quantized token dictionary, constructed by pre-training a Vector Quantized-VAE (VQ-VAE)~\cite{van2017vqvae} on motion capture datasets~\cite{mahmood2019amass, moyo}, to mitigate 3D pose ambiguity. CameraHMR~\cite{patel2024camerahmr} improves the 4DHuman~\cite{goel20234dhuman} dataset through CamSMPLify, which refines body representations using dense keypoints to overcome the average body limitation. To tackle the fixed-camera constraint, it introduces HumanFOV~\cite{patel2024camerahmr}, transforming weak-perspective projection into a fully perspective projection.

\paragraph{SMPL-to-SKEL dataset conversion.}
The first large-scale 3D pseudo-ground-truth (p-GT) dataset was introduced by 4DHuman~\cite{goel20234dhuman}, which extends to unlabeled datasets such as InstaVariety~\cite{kanazawa2019insta}, AVA~\cite{gu2018ava}, and AI Challenger~\cite{sun2019aic} by generating p-GT annotations. For each image, an off-the-shelf detector~\cite{detr} and a body keypoint estimator are applied to obtain bounding boxes and 2D keypoints. A SMPL~\cite{loper2015smpl} mesh is then fitted to these keypoints using ProHMR~\cite{prohmr}, producing pseudo-ground-truth SMPL parameters. Despite its scale and utility, the 4DHuman dataset still contains inaccuracies and artifacts. To that end, CameraHMR~\cite{patel2024camerahmr} refines the dataset using its proposed CamSMPLify method, alongside additional techniques to enhance annotation quality. While large-scale datasets of SMPL parameters have been constructed, there remains a need to develop datasets for SKEL~\cite{SKEL} parameters. HSMR~\cite{xia2025hsmr} addresses this by first converting the 4DHuman SMPL parameters to SKEL parameters using SKEL fitting~\cite{SKEL}. To mitigate the inaccuracies in 4DHuman annotations, we further apply SKEL fitting to the refined SMPL annotations provided by CameraHMR~\cite{patel2024camerahmr}.

\section{Method}

\begin{figure*}[t]
    \centering
    \includegraphics[width=\textwidth]{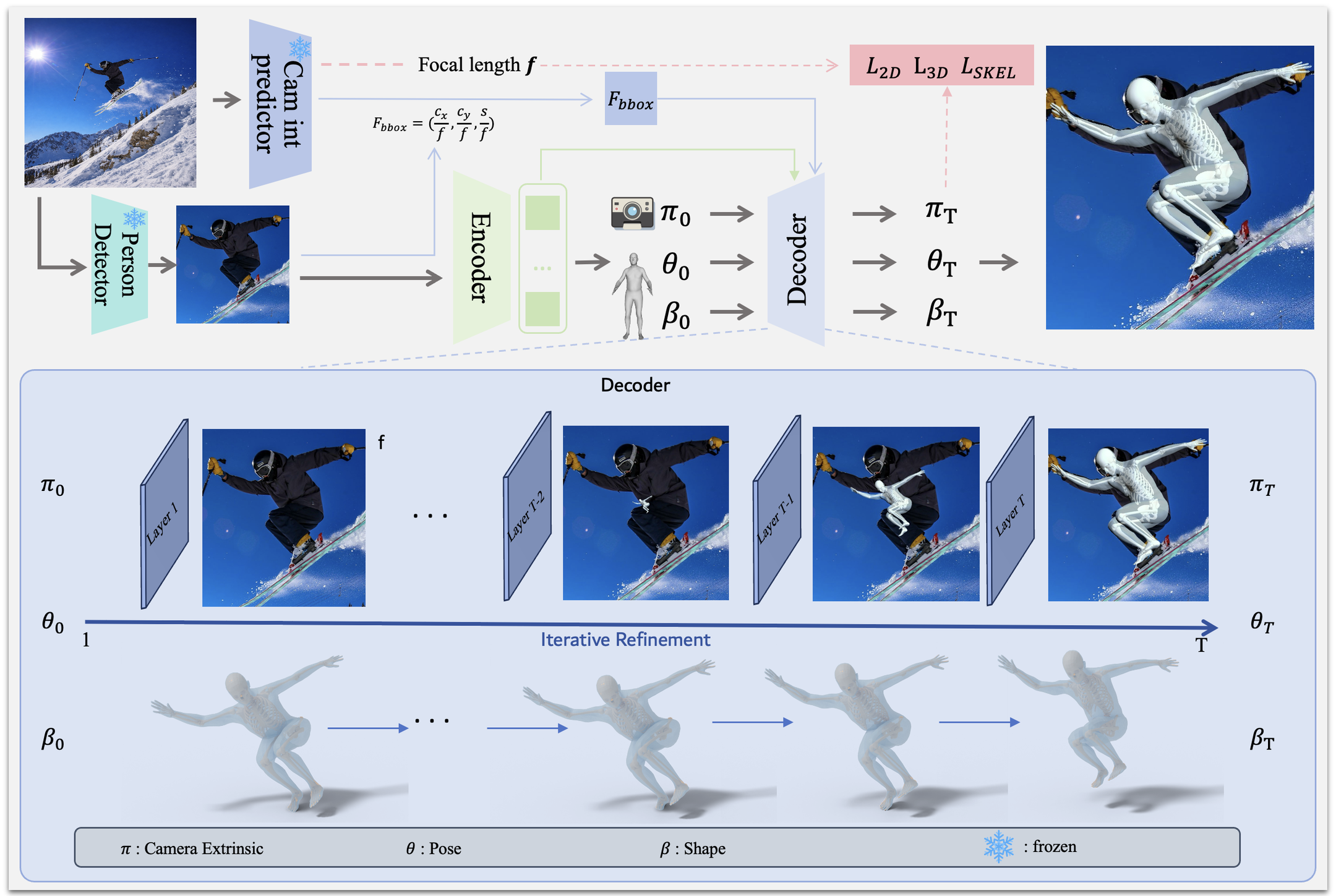}
    \caption{\textbf{Overview of the proposed SKEL-CF.} Our method estimates SKEL parameters from a single image using an encoder–decoder architecture that performs coarse-to-fine estimation. The encoder produces initial predictions of the \textit{camera extrinsics} $\boldsymbol{\pi}$, \textit{shape parameters} $\boldsymbol{\beta}$, and \textit{pose parameters} $\boldsymbol{\theta}$. The decoder then refines these predictions iteratively across layers. In addition, we adopt the pretrained camera model from CameraHMR~\cite{patel2024camerahmr} to estimate the focal length, and keep its parameters frozen throughout training.}
    \label{fig:overview}
\end{figure*}

In this section, we introduce our approach, SKEL-CF, whose overall pipeline is illustrated in Fig.~\ref{fig:overview}. SKEL-CF aims to estimate SKEL parameters from a single image using an encoder–decoder architecture that performs coarse-to-fine estimation. The encoder produces initial predictions of the camera extrinsics $\boldsymbol{\pi}_0$, shape parameters $\boldsymbol{\beta}_0$, and pose parameters $\boldsymbol{\theta}_0$, while the decoder iteratively refines these predictions across layers. To enhance robustness to camera variations, we adopt the pretrained camera intrinsic predictor from CameraHMR~\cite{patel2024camerahmr} to estimate the focal
length, keeping its parameters frozen throughout training.
This section is organized as follows. We first introduce the SKEL model and the HMR-SKEL dataset that we construct to supervise SKEL learning in Section~\ref{subsec:preliminary}. We then describe the details of SKEL-CF and its implementation in Section~\ref{subsec:skel_cf}.

\subsection{Preliminaries}
\label{subsec:preliminary}

\paragraph{SKEL model.}
The SKEL model~\cite{SKEL} is a parametric representation of the human body that enables the unified reconstruction of both the surface mesh and the skeleton mesh. Given the pose parameters $q \in \mathbb{R}^{46}$ and the shape parameters $\boldsymbol{\beta} \in \mathbb{R}^{10}$ (corresponding to the top 10 shape components), SKEL generates the human mesh $M \in \mathbb{R}^{6890 \times 3}$ and the skeleton mesh $S_{k} \in \mathbb{R}^{24752 \times 3}$.
Unlike SMPL, which represents pose parameters $\theta \in \mathbb{R}^{72}$ using unconstrained three-degree-of-freedom ball joints, SKEL constrains the motion of each joint. This reduces the pose parameter space from 72 to 46. For example, SKEL models the elbow as a hinge joint by limiting its degrees of freedom. This formulation ensures anatomically consistent and biomechanically valid poses~\cite{SKEL,xia2025hsmr}. Consequently, SKEL produces physically plausible human motion and is particularly suitable for downstream applications such as motion analysis, rehabilitation, and human–robot interaction.

\paragraph{HMR-SKEL dataset.} HSMR~\cite{xia2025hsmr} generated pseudo ground-truth (p-GT) SKEL parameters by converting SMPL parameters into the SKEL space. While this provided a practical starting point, its quality was limited by low-resolution images and noisy SMPL annotations in the original dataset. With the release of refined SMPL annotations from CameraHMR~\cite{patel2024camerahmr}, we revisit this conversion process to obtain higher-quality SKEL pseudo ground truth. Following the HSMR setup, which is based on HMR2.0~\cite{goel20234dhuman}, we use data from Human3.6M~\cite{human36m}, MPI-INF-3DHP~\cite{mpi-inf}, COCO~\cite{coco}, MPII~\cite{mpii}, AI Challenger~\cite{aic}, and InstaVariety~\cite{insta}. We exclude AVA because CameraHMR does not provide refined SMPL annotations for this subset. We denote the resulting dataset as HMR-SKEL.



Our optimization pipeline follows the fitting protocol of SKEL~\cite{SKEL}. We first reconstruct a reference human mesh from the high-quality annotations provided by CameraHMR~\cite{patel2024camerahmr}, which is used as the target SMPL mesh. Given an initial set of SKEL parameters, we treat them as learnable variables and feed them into the SKEL model to generate a SKEL mesh. We then compute the alignment loss between the generated SKEL mesh and the target SMPL mesh, and use this loss to iteratively update the SKEL parameters. To improve optimization stability, we perform the fitting in a hierarchical manner: we first optimize the lower-body parameters, then the upper-body parameters, and finally the full-body parameters. Moreover, since both SMPL and SKEL share the same global orientation and translation terms, we keep these variables fixed during optimization; empirically, allowing them to vary leads to worse alignment. The full process takes about 58 hours to process 3M images on a single RTX 3090 GPU.

\subsection{Our method: SKEL-CF} \label{subsec:skel_cf}

\paragraph{Overall pipeline.} 
As illustrated in Fig.~\ref{fig:overview}, SKEL-CF adopts an encoder–decoder architecture with a coarse-to-fine strategy to estimate SKEL parameters from a single RGB image. The target parameter set is defined as $\boldsymbol{\Theta} = \{\boldsymbol{\theta}, \boldsymbol{\beta}, \boldsymbol{\pi}\}$, where $\boldsymbol{\theta}$ denotes human pose, $\boldsymbol{\beta}$ is body shape, and $\boldsymbol{\pi}$ indicates camera extrinsics. Given an input image, following previous work~\cite{patel2024camerahmr}, we first detect the person using an off-the-shelf human detector and convert the cropped region into visual tokens. We also estimate the camera intrinsic parameter, specifically the focal length $f$, from the full image using a pretrained predictor from CameraHMR~\cite{patel2024camerahmr}, which remains frozen during training. The visual tokens are passed through the encoder to produce initial predictions $\boldsymbol{\Theta}_0 = \{\boldsymbol{\theta}_0, \boldsymbol{\beta}_0, \boldsymbol{\pi}_0\}$ along with contextual visual features. These predictions are then progressively refined through $T$ decoder layers. At each stage, refinement is guided by the encoded visual features and geometric cues $\mathcal{F}_{\text{bbox}} = \left(\frac{c_x}{f}, \frac{c_y}{f}, \frac{s}{f}\right)$, where $(c_x, c_y)$ and $s$ denote the bounding box center and scale. After $T$ refinement steps, the final output is $\boldsymbol{\Theta}_T = \{\boldsymbol{\theta}_T, \boldsymbol{\beta}_T, \boldsymbol{\pi}_T\}$, yielding progressively improved SKEL parameter estimates. The focal-normalized feature $\mathcal{F}_{\text{bbox}}$ is injected into the decoder as bounding-box geometry, where it guides projection-sensitive residual corrections after the image crop has been encoded.

\paragraph{Learning objectives.} 
Denoting the target parameters as
$\boldsymbol{\hat{\Theta}} = \{\boldsymbol{\hat{\theta}}, \boldsymbol{\hat{\beta}}, \boldsymbol{\hat{\pi}}\}$, the learning objective integrates keypoint-level and parameter-level supervision:
\begin{equation}
\mathcal{L}_{\text{skel}} (\boldsymbol{\Theta}, \boldsymbol{\hat{\Theta}}) = 
\underbrace{\lambda_{kp} \mathcal{L}_{\text{kp}}}_{\text{keypoint-level}} 
+ 
\underbrace{\lambda_{\beta} \mathcal{L}_{\beta} + \lambda_{\theta} \mathcal{L}_{\theta}}_{\text{parameter-level}}.
\end{equation}
where $\lambda_{kp}$, $\lambda_{\beta}$, and $\lambda_{\theta}$ are hyperparameters controlling the relative weight of each term. 
This formulation allows us to simultaneously supervise the reconstructed 3D/2D joints and the underlying pose and shape parameters. Note that no direct camera extrinsic supervision $\boldsymbol{\hat{\pi}}$ is provided; the ability to predict camera extrinsics is only implicitly learned through 2D keypoint reprojection, making the training challenging.
\paragraph{Keypoint-level supervision.} 
Given the pose $\boldsymbol{\theta}$ and shape $\boldsymbol{\beta}$, we first reconstruct 3D joint positions 
$\mathbf{J}_{3d} \in \mathbb{R}^{K \times 3}$ via forward kinematics, where $K$ is the number of joints. 
These 3D joints are projected onto the image plane using the predicted camera parameters to obtain 2D joint locations 
$\mathbf{J}_{2d} \in \mathbb{R}^{K \times 2}$. 
The coarse keypoint loss $\mathcal{L}_{\text{kp}}$ enforces consistency with pseudo-ground-truth 3D and 2D joints:
\begin{equation}
\mathcal{L}_{\text{kp}} = \| \hat{\mathbf{J}}_{3d} - \mathbf{J}_{3d} \|_1 + \| \hat{\mathbf{J}}_{2d} - \mathbf{J}_{2d} \|_1.
\end{equation}

\paragraph{Parameter-level supervision.} 
Since the proposed dataset HMR-SKEL provides reliable pose $\hat{\boldsymbol{\theta}}$ and shape $\hat{\boldsymbol{\beta}}$ parameters, we 
impose supervision directly on these values:
\begin{equation}
\mathcal{L}_{\beta} = \| \hat{\boldsymbol{\beta}} - \boldsymbol{\beta} \|_1, 
\quad
\mathcal{L}_{\theta} = \| \hat{\boldsymbol{\theta}} - \boldsymbol{\theta} \|_1.
\end{equation}

\paragraph{Coarse-to-fine refinement.}
SKEL-CF adopts a coarse-to-fine refinement strategy.
The encoder first produces an initial estimate of the parameters, denoted as $\boldsymbol{\Theta}_0 = \{\boldsymbol{\theta}_0, \boldsymbol{\beta}_0, \boldsymbol{\pi}_0\}$. The decoder then refines these predictions across $T$ layers, where the $i$-th layer outputs $\boldsymbol{\Theta}_i = \{\boldsymbol{\theta}_i, \boldsymbol{\beta}_i, \boldsymbol{\pi}_i\}$. The final layer produces the refined prediction $\boldsymbol{\Theta}_T$. The encoder prediction and the final decoder prediction are supervised using the same objective function $\mathcal{L}_{\text{skel}}$:
\begin{equation}
\mathcal{L}_{\text{enc}} = \mathcal{L}_{\text{skel}} (\boldsymbol{\Theta}_0, \boldsymbol{\hat{\Theta}}), 
\quad 
\mathcal{L}_{\text{dec}} = \mathcal{L}_{\text{skel}} (\boldsymbol{\Theta}_T, \boldsymbol{\hat{\Theta}})
\end{equation}
\paragraph{Iterative refinement.}

We use iterative refinement with layer-wise supervision. The refinement process consists of two operations: coarse-to-fine initialization and residual refinement. The encoder first predicts an image-dependent coarse estimate $\boldsymbol{\Theta}_0$, and the decoder then updates this estimate \textit{layer by layer} by predicting residual corrections. To supervise the intermediate decoder layers efficiently, we apply an auxiliary pose-parameter loss:
\begin{equation}
\mathcal{L}_{\text{refine}} = 
\sum_{i=1}^{T-1}
\| \hat{\boldsymbol{\theta}}_i - \boldsymbol{\theta}_i \|_1,
\end{equation}
where $T$ denotes the total number of decoder layers, and $\hat{\boldsymbol{\theta}}_i$ denotes the corresponding ground-truth pose parameters. We observe that optimizing $\mathcal{L}_{\text{skel}}$ at intermediate decoder layers achieves performance comparable to $\mathcal{L}_{\text{refine}}$. However, as computing keypoint-level predictions through the SKEL forward process is computationally expensive, we limit the optimization to the pose parameters for efficiency.

Note that coarse-to-fine strategy and iterative refinement are widely adopted in the DETR-based object detection frameworks~\cite{zhu2020deformable,detr}. The experiments to demonstrate the effectiveness of iterative refinement are provided in the supplementary material.

\paragraph{Overall optimization goal.}
The total training objective $\mathcal{L}_{\text{tot}}$ aggregates the encoder and decoder losses, together with the refinement term:

\begin{align}
\mathcal{L}_{\text{tot}} = \mathcal{L}_{\text{dec}} +  \mathcal{L}_{\text{enc}} + \lambda_{ref} \mathcal{L}_{\text{refine}},
\end{align}
where $\lambda_{ref}$ is the coefficient for the refinement loss.

\paragraph{Implementation details.}

Following HSMR~\cite{xia2025hsmr}, we adopt ViTPose-H~\cite{xu2022vitpose}, pretrained on COCO~\cite{coco}, as the encoder backbone. ViTPose-H consists of 32 transformer layers, each with 16 attention heads and a hidden dimension of 1280. The decoder is a standard transformer decoder with $L = 6$ layers, consistent with the design in HSMR~\cite{xia2025hsmr}. We train SKEL-CF using the AdamW~\cite{loshchilov2017decoupled} optimizer with $\beta_1$, $\beta_2$ setting to 0.9, 0.999, weight decay setting to $1 \times10^{-4}$, a batch size of 64 and a learning rate of $1\times10^{-5}$, preceded by a one-epoch warm-up. Training is conducted for 30 epochs on 8 NVIDIA A100 GPUs, taking approximately 120 hours in total. The hyper-parameters are set as $\lambda_{kp}=0.05$, $\lambda_{\beta}=0.0005$, $\lambda_{\theta}=0.001$, and $\lambda_{ref}=0.1$. Notably, our training schedule is significantly shorter than that of HSMR~\cite{xia2025hsmr}, which trains for 100 epochs. 


\section{Experiment}
In this section, we first introduce the datasets and evaluation metrics in Section~\ref{subsec:metrics}. We then provide a comparison between SKEL-based methods in Section~\ref{subsec:SKEL} and SMPL-based approaches in Section~\ref{subsec:SMPL}. Finally, we present the ablation studies and additional discussions in Section~\ref{subsec:discussion}.

\subsection{Datasets and evaluation metrics} 
\label{subsec:metrics}

\paragraph{Evaluated datasets.}
We evaluate SKEL-CF on five representative datasets covering diverse environments and motion complexities, following standard practice~\cite{patel2024camerahmr,xia2025hsmr,kanazawa2018hmr,dwivedi2024tokenhmr}.
\textit{3DPW}~\cite{3dpw} provides in-the-wild videos with accurate 3D annotations captured by moving cameras.
\textit{Human3.6M}~\cite{human36m} includes controlled indoor actions (e.g., sitting, walking, greeting) with motion-capture ground truth.
\textit{EMDB}~\cite{kaufmann2023emdb} offers high-quality pose and shape sequences captured with electromagnetic sensors and handheld cameras.
\textit{SPEC-SYN}~\cite{spec-syn} is a synthetic dataset with diverse camera variations, enabling controlled evaluation under varying camera settings. In contrast, \textit{MOYO}~\cite{moyo} contains large-scale yoga sequences with extreme poses, frequent self-occlusions, and ground contact, providing the most challenging and diverse test scenario. 

Since most yoga videos in MOYO starts and ends with the same static poses, we select the front-view camera among the eight available viewpoints and remove the first and last 25\% of frames to retain the complex and diverse motion segments, forming the curated \textit{MOYO-HARD} subset. More details about MOYO-HARD are provided in the supplementary material, where we include visualizations of the removed first and last 25\% segments of the MOYO videos.

\paragraph{Evaluation metrics.} 
We evaluate the 3D human body recovery using MPJPE, PA-MPJPE, and PVE.
Among them, MPJPE and PA-MPJPE serve as sparse evaluations, measuring the Euclidean distance between the predicted and ground-truth kinematic joints before and after rigid alignment, respectively.
In contrast, PVE provides a dense evaluation, computing the vertex-level error between the predicted and ground-truth meshes, thereby offering a more comprehensive assessment of surface reconstruction accuracy. The 2D keypoint metrics, PCK@0.05 and PCK@0.1, which quantify the ratio of projected keypoints within a normalized distance threshold to the ground truth.

\begin{table*}[t]
    \caption{
    \textbf{Quantitative comparison of skeleton and surface mesh recovery using the SKEL~\cite{SKEL} human model on 3DPW~\cite{3dpw}, Human3.6M~\cite{human36m}, and MOYO~\cite{moyo}.} The proposed SKEL-CF delivers consistent accuracy gains over previous end-to-end approach. $^{*}$ denotes results reported by~\cite{xia2025hsmr}, $^{\diamond}$ denotes fitting the outputs of~\cite{patel2024camerahmr} to the SKEL~\cite{SKEL} model.
    }
  \centering
  \footnotesize
\resizebox{\textwidth}{!}{
    \begin{tabular}{l|cc|cc|cc|cc}
    \toprule
    \multirow{2}{*}{Methods} &
    \multicolumn{2}{c|}{3DPW~\cite{3dpw}} &
    \multicolumn{2}{c|}{Human3.6M~\cite{human36m}} &
    \multicolumn{2}{c|}{MOYO~\cite{moyo}} &
    \multicolumn{2}{c}{MOYO-HARD~}\\
    \cmidrule(lr){2-9}
    & MPJPE~$\downarrow$ & PA-MPJPE~$\downarrow$ &MPJPE~$\downarrow$ & PA-MPJPE~$\downarrow$& MPJPE~$\downarrow$ & PA-MPJPE~$\downarrow$& MPJPE~$\downarrow$ & PA-MPJPE~$\downarrow$ \\
    \midrule
    \multicolumn{9}{c}{\textbf{Two-stage Approaches}} \\
    \rowcolor{LightGray}
    HMR2.0 + SKEL fit~\cite{xia2025hsmr,goel20234dhuman}$^{*}$ & 81.0 & 54.4  & 53.6 & 34.1 & 130.5 & 93.7 & - & -\\
    \rowcolor{LightGray}
    CameraHMR + SKEL fit~\cite{xia2025hsmr, patel2024camerahmr}$^{\diamond}$ & 70.4 & 41.8  & - & - & \textbf{75.5} & \textbf{49.9} & \textbf{88.7} & \textbf{61.4}\\
    \multicolumn{9}{c}{\textbf{End-to-end Approaches}} \\
    HSMR~\cite{xia2025hsmr} & 81.5 & 54.8  &  50.4 & 32.9 & 104.5 & 79.6   & 120.0 & 97.7\\
    \textbf{SKEL-CF (Ours)} &
    \textbf{61.5} &
    \textbf{38.7}   &
    \textbf{39.0}  &
    \textbf{31.2}  &
    85.0  &
    51.4  &
    90.0  &
    61.5  \\
    \bottomrule
    \end{tabular}
}
  \label{tab:skel_recon}
\end{table*}

\begin{figure}[!tp]
    \centering
    \includegraphics[width=\textwidth]{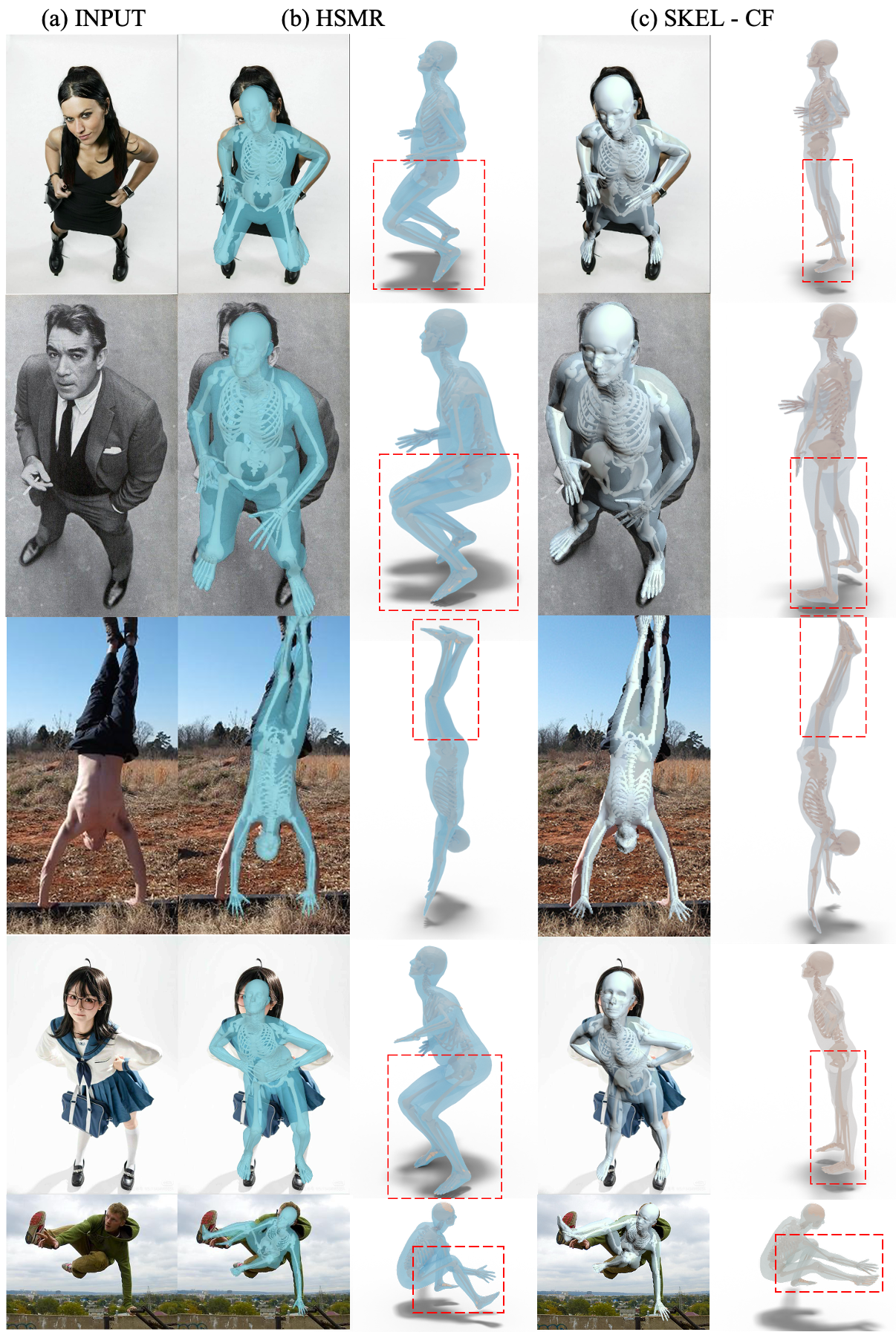}
    \caption{\textbf{Visual comparison between the proposed SKEL-CF and HSMR~\cite{xia2025hsmr}.} Our proposed SKEL-CF achieves more precise skeletal estimations (best viewed in zoom). Additional visual results are provided in the supplementary material.}
    \label{fig:compare-hsmr}
\end{figure}

\subsection{Comparison with SKEL-based approaches.}
\label{subsec:SKEL}
\paragraph{Quantitative results.}
We compare SKEL-based methods in Table~\ref{tab:skel_recon} and Table~\ref{tab:smpl_recon_humanmodel}. Among end-to-end approaches, SKEL-CF consistently outperforms prior work HSMR~\cite{xia2025hsmr} across all datasets. On the challenging MOYO~\cite{moyo} benchmark, SKEL-CF achieves up to 18.6\% MPJPE and 35.4\% PA-MPJPE improvements, demonstrating strong robustness under large pose variations and occlusions. We also evaluate two two-stage baselines that first estimate SMPL parameters and then fit them to SKEL: HMR2.0 + SKEL fitting and CameraHMR + SKEL fitting. While CameraHMR + SKEL fit achieves competitive results on MOYO and MOYO-HARD, its performance drops on 3DPW, indicating limited stability across datasets. Moreover, the fitting-based approach is computationally expensive (approximately 4 minute per image on RTX 3090), whereas SKEL-CF runs in a single forward pass (approximately 0.8 second per image), making it both more efficient and more reliable.

\paragraph{Visual results.} We provide a visual comparison with HSMR~\cite{xia2025hsmr} in Fig.~\ref{fig:compare-hsmr}. As shown, the proposed SKEL-CF achieves more accurate skeleton and surface reconstruction. More visual examples similar to Fig.~\ref{fig:compare-hsmr} are included in supplementary material. 

\begin{table}[t]
\caption{\textbf{Quantitative comparison of mesh recovery using the SMPL~\cite{loper2015smpl} and SKEL\cite{keller2023skin}} human model on 3DPW~\cite{3dpw}, EMDB~\cite{kaufmann2023emdb}, and SPEC-SYN~\cite{spec-syn}.The complete comparison including MOYO and MOYO-HARD is provided in Table~\ref{tab:supp_smpl_recon_full} of the supplementary material.}
\centering
\scriptsize
\setlength{\tabcolsep}{2pt}
\resizebox{\columnwidth}{!}{
\begin{tabular}{l|ccc|ccc|ccc}
\toprule
\multirow{2}{*}{Method} &
\multicolumn{3}{c|}{3DPW~\cite{3dpw}} &
\multicolumn{3}{c|}{EMDB~\cite{kaufmann2023emdb}} &
\multicolumn{3}{c}{SPEC-SYN~\cite{spec-syn}} \\
\cmidrule(lr){2-4}
\cmidrule(lr){5-7}
\cmidrule(lr){8-10}
&
MPJPE$\downarrow$ & PA-MPJPE$\downarrow$ & PVE$\downarrow$ &
MPJPE$\downarrow$ & PA-MPJPE$\downarrow$ & PVE$\downarrow$ &
MPJPE$\downarrow$ & PA-MPJPE$\downarrow$ & PVE$\downarrow$ \\
\midrule

\multicolumn{10}{c}{\textbf{SMPL-based Approaches}} \\

\rowcolor{LightGray}
SPEC~\cite{spec-syn} & 96.5 & 53.2 & 118.5 & 138.9 & 87.7 & 161.3 & 83.5 & 56.9 & 98.9 \\
\rowcolor{LightGray}
CLIFF~\cite{li2022cliff} & 69.0 & 43.0 & 81.2 & 103.5 & 68.3 & 123.7 & 128.5 & 55.8 & 139.0 \\
\rowcolor{LightGray}
HMR2.0a~\cite{goel20234dhuman} & 69.8 & 44.4 & 82.2 & 97.8 & 61.5 & 120.0 & 133.3 & 55.8 & 153.0 \\
\rowcolor{LightGray}
TokenHMR~\cite{dwivedi2024tokenhmr} & 70.5 & 43.8 & 86.0 & 88.1 & 49.8 & 104.2 & 110.5 & 51.8 & 127.6 \\
\rowcolor{LightGray}
WHAM~\cite{shin2024wham} & 57.8 & 35.9 & 68.7 & 79.7 & 50.4 & 94.4 & - & - & - \\
\rowcolor{LightGray}
ReFit~\cite{wang2023refit} & 57.6 & 38.2 & 67.6 & 91.7 & 55.5 & 106.2 & 103.6 & 51.3 & 116.3 \\
\rowcolor{LightGray}
CLIFF~\cite{li2022cliff} & 72.0 & 46.6 & 85.0 & 97.1 & 61.3 & 113.2 & 109.9 & 55.6 & 124.6 \\
\rowcolor{LightGray}
HMR2.0b~\cite{goel20234dhuman} & 81.3 & 54.3 & 93.1 & 118.5 & 79.2 & 140.6 & 150.7 & 67.6 & 172.9 \\
\rowcolor{LightGray}
CameraHMR~\cite{patel2024camerahmr} & 62.7 & 38.7 & \textbf{73.4} & 73.2 & \textbf{43.9} & 85.6 & \textbf{66.0} & \textbf{37.0} & \textbf{79.1} \\

\midrule
\multicolumn{10}{c}{\textbf{SKEL-based Approaches}} \\

HSMR~\cite{xia2025hsmr} & 81.5 & 54.8 & - & - & - & - & - & - & - \\
SKEL-CF (Ours) & \textbf{61.5} & \textbf{38.7} & 73.5 & \textbf{72.0} & 44.5 & \textbf{84.7} & 69.4 & 37.1 & 83.4 \\

\bottomrule
\end{tabular}
}
\label{tab:smpl_recon_humanmodel}
\end{table}

\begin{figure}[!tp]
    \centering
    \includegraphics[width=\textwidth]{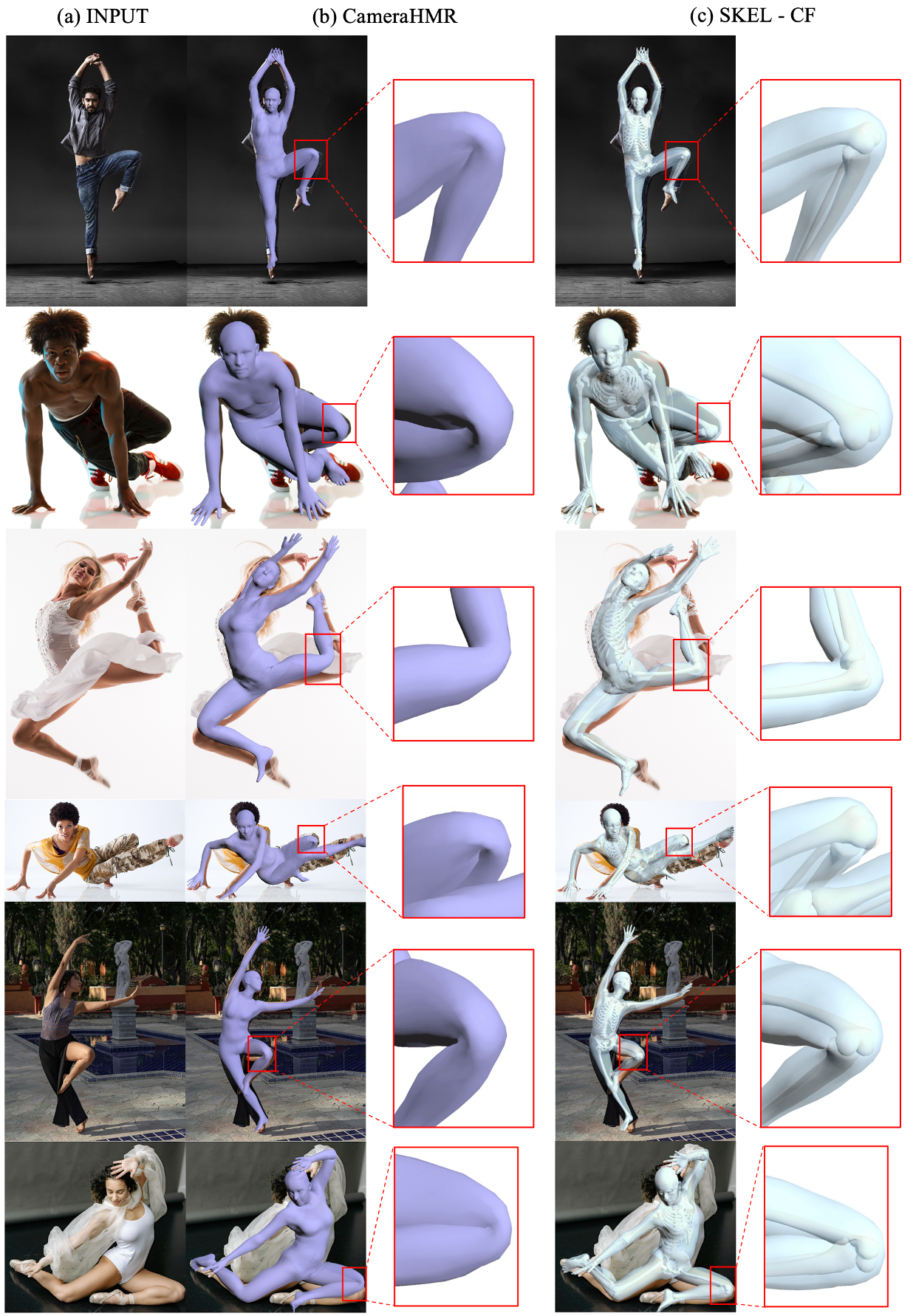}
    \caption{\textbf{Visual comparison between the proposed SKEL-CF and CameraHMR~\cite{patel2024camerahmr}.} The proposed SKEL-CF is built upon the SKEL~\cite{SKEL} representation, which enforces anatomically consistent joint motion, resulting in more natural pose predictions compared to the SMPL~\cite{loper2015smpl}-based CameraHMR~\cite{patel2024camerahmr} (best viewed in zoom). Additional visual examples are provided in the supplementary material.}
    \label{fig:ompare-camhmr}
\end{figure}

\subsection{Comparison with SMPL-based approaches}
\label{subsec:SMPL}
\paragraph{Quantitative results.} We further compare SKEL-CF with recent state-of-the-art SMPL-based human mesh recovery methods across four representative datasets: 3DPW~\cite{3dpw}, EMDB~\cite{kaufmann2023emdb}, SPEC-SYN~\cite{spec-syn}, MOYO~\cite{moyo} and more challenging MOYO-HARD, as summarized in Table~\ref{tab:smpl_recon_humanmodel}. Note that SMPL~\cite{loper2015smpl} is less constrained compared to SKEL~\cite{SKEL}, enabling strong quantitative performance but occasionally producing anatomical inconsistency or unnatural joint motions(see Fig.~\ref{fig:teaser}). Despite this inherent difference, SKEL-CF achieves comparable performance to CameraHMR~\cite{patel2024camerahmr}, the strongest SMPL-based model trained on 4DHuman dataset, demonstrating that kinematically constrained SKEL representations can reach a similar level of numerical precision while maintaining stronger physical plausibility. Note that, the the 2D keypoint results are provided in the supplementary material. 



\paragraph{Visual results.} We provide a visual comparison with the SMPL~\cite{loper2015smpl}-based state-of-the-art method CameraHMR~\cite{patel2024camerahmr} in Fig.~\ref{fig:ompare-camhmr}. As shown, the proposed SKEL-CF produces anatomically more consistent joint motions, benefiting from the structural constraints of the SKEL~\cite{SKEL} model. In contrast, CameraHMR tends to generate unnatural joint configurations under complex motion scenarios. Additional visualizations similar to Fig.~\ref{fig:ompare-camhmr} are provided in supplementary material. 

\subsection{Discussion}
\label{subsec:discussion}


\begin{table*}[t]
\caption{\textbf{Ablation study on MOYO-HARD and COCO.}
\textit{Cam} denotes camera intrinsic modeling, \textit{C2F} denotes coarse-to-fine estimation, and \textit{Refine} denotes iterative  refinement.}
\centering
\resizebox{\textwidth}{!}{
\setlength{\tabcolsep}{2pt}
\renewcommand{\arraystretch}{0.9}
\small
\begin{threeparttable}
\begin{tabular}{l|cccc|ccc|cc}
\toprule
\multirow{2}{*}{Method} &
\multicolumn{4}{c|}{Components} &
\multicolumn{3}{c|}{MOYO-HARD} &
\multicolumn{2}{c}{COCO~\cite{coco}} \\
\cmidrule(lr){2-5}
\cmidrule(lr){6-8}
\cmidrule(lr){9-10}
& Cam & C2F & Refine & Dataset
& MPJPE$\downarrow$
& PA-MPJPE$\downarrow$
& PVE$\downarrow$
& PCK@0.05$\uparrow$
& PCK@0.1$\uparrow$ \\
\midrule
\rowcolor{LightGray}
Baseline (HSMR~\cite{xia2025hsmr}) &
\xmark & \xmark & \xmark &
HMR2.0 + SKELify &
120.0 & 97.7 & 140.5 &
\textbf{0.86} & \textbf{0.96} \\

Baseline w. HMR-SKEL &
\xmark & \xmark & \xmark &
HMR-SKEL &
103.6 & 67.4 & 121.4 &
0.76 & 0.91 \\

Ours w.o Cam &
\xmark & \cmark & \cmark &
HMR-SKEL &
98.8 & 66.1 & 113.7 &
0.79 & 0.93 \\

Ours w.o C2F &
\cmark & \xmark & \cmark &
HMR-SKEL &
91.5 & 63.1 & 105.6 &
0.67 & 0.91 \\

Ours w.o Refine &
\cmark & \cmark & \xmark &
HMR-SKEL &
92.7 & 65.4 & 107.7 &
0.77 & 0.92 \\

Only Cam &
\cmark & \xmark & \xmark &
HMR-SKEL &
92.7 & 66.4 & 107.4 &
0.77 & 0.92 \\

\textbf{Ours} &
\cmark & \cmark & \cmark &
HMR-SKEL &
\textbf{90.0} & \textbf{61.5} & \textbf{102.5} &
0.80 & 0.93 \\

\bottomrule
\end{tabular}
\end{threeparttable}
\label{tab:moyo_ablation}}
\end{table*}

\begin{figure*}[t]
    \centering
    \includegraphics[width=\textwidth]{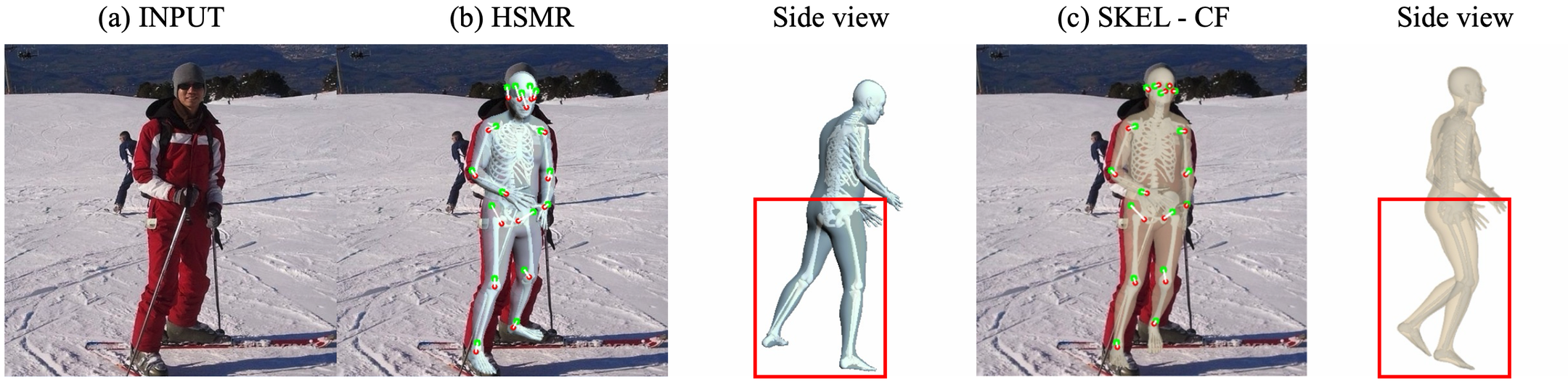}
    \caption{\textbf{COCO 2D metric saturation.} Similar 2D PCK can hide large 3D pose differences. \textcolor{red}{Red} and \textcolor{green}{green} denote predicted keypoints and ground-truth 2D keypoints, respectively.}
    \label{fig:rebuttal_coco_2d_}
\end{figure*}
\paragraph{Ablation study.} 
To analyze the contribution of each component, we conduct controlled ablations on MOYO-HARD, and COCO~\cite{coco}. Note that more ablation on 3DPW~\cite{3dpw}, MOYO~\cite{moyo} are provided in the supplementary material. As shown in Table~\ref{tab:moyo_ablation}. We start from an HSMR-style baseline that disables the camera intrinsic predictor, coarse-to-fine (C2F) initialization, and iterative refinement. Comparing HSMR with the same architecture trained on HMR-SKEL isolates the effect of the improved annotations: on MOYO, PA-MPJPE improves from 79.6 to 53.7. Building on this same-data baseline, the full SKEL-CF pipeline further improves MOYO PA-MPJPE from 53.7 to 51.4, showing additional gains beyond the dataset. C2F and refinement bring more visible gains on the harder MOYO-HARD subset, where the full model improves PVE over the Cam-only variant from 107.4 to 102.5, and also recovers stronger COCO 2D alignment than the without C2F or without iterative refinement variants. Note that COCO PCK evaluates only the 2D projection of keypoints and can be insensitive to 3D pose errors. Fig.~\ref{fig:rebuttal_coco_2d_} shows that similar 2D PCK can correspond to noticeably different 3D reconstructions, so we treat COCO PCK as a complementary projection metric rather than the primary evidence for SKEL recovery quality.

\begin{figure}[t]
    \centering
    \includegraphics[width=0.95\linewidth]{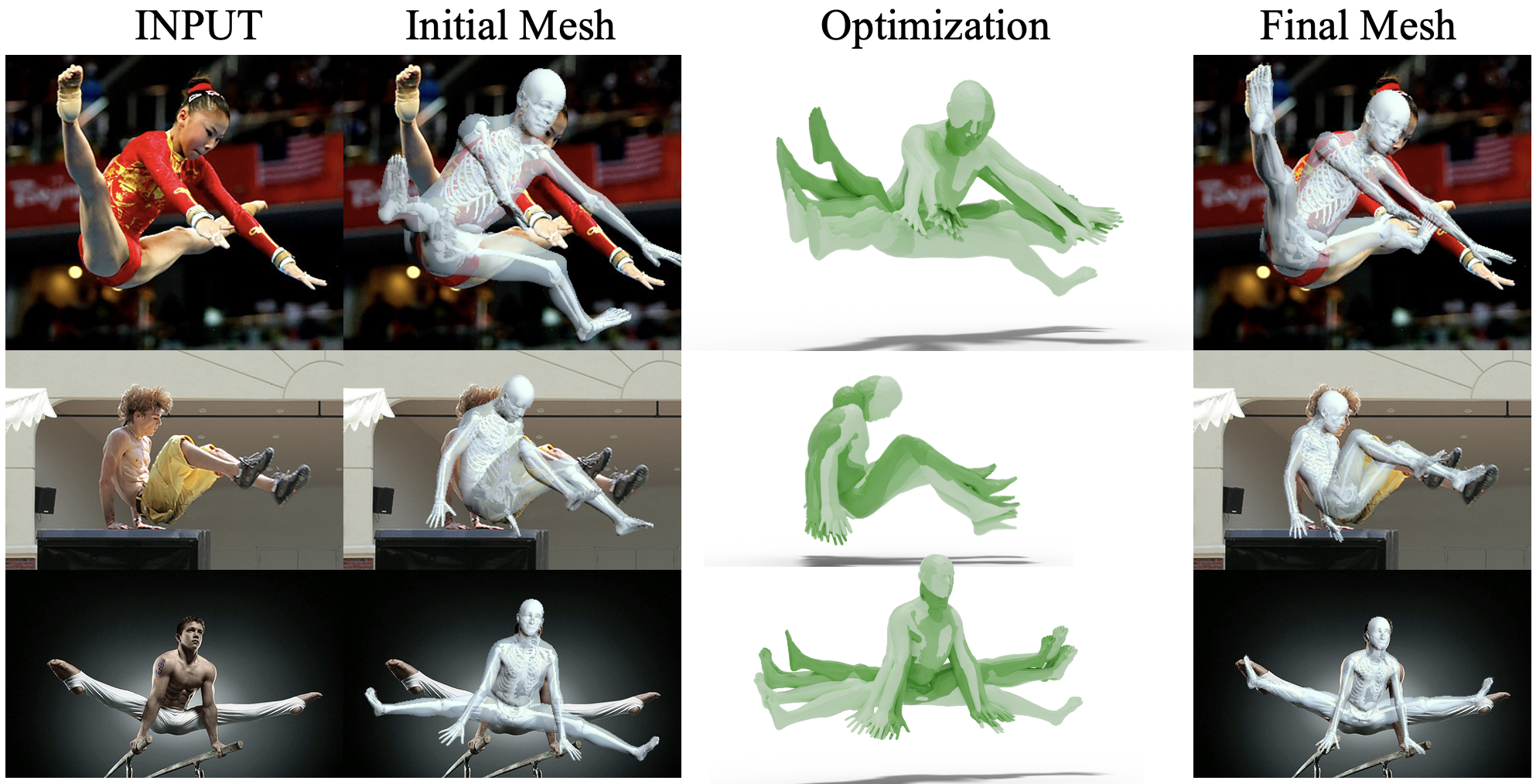}
    \caption{\textbf{Illustration of the mesh refinement.} The light green meshes represent the initial coarse estimations, while the dark green meshes denote the iteratively refined final results.}
    \label{fig:processive}
\end{figure}

\paragraph{Iterative mesh refinement.} We visualize the iterative mesh refinement process of SKEL-CF in Fig.~\ref{fig:processive}, which clearly illustrates the effectiveness of the proposed coarse-to-fine strategy. Additional qualitative results are provided in the supplementary material. Quantitative validation is reported in Table~\ref{tab:pred_compare}. We further investigate two variants: \textit{i)} applying the full supervision loss $L_{\text{skel}}$ to intermediate decoder layers (\textit{Full super.}), instead of using only the refinement loss $L_{\text{refine}}$ as sparse supervision; \textit{ii)} replacing the six-layer decoder with a single decoder layer that is reused six times~\cite{lan2019albert} (\textit{Looped Decoder}). Results show that sparse supervision achieves performance comparable to full supervision. However, computing $\mathcal{L}_{\text{skel}}$ is significantly more expensive, as it requires forwarding the SKEL model to obtain the keypoint loss. Therefore, sparse supervision is more computationally efficient and preferable in practice. In contrast, replacing the multi-layer decoder with a single shared layer substantially reduces the decoder’s representational capacity, leading to inferior reconstruction quality and weaker generalization.
\begin{table}[t]
\caption{\textbf{Ablation study of iterative refinement.} \textit{Sparse super.} applies refinement supervision across decoder layers, \textit{Full super.} indicates supervision with full SKEL loss $\mathcal{L}_{skel}$ at every layer, and \textit{Looped Decoder} repeats a single layer for multiple refinement steps. ``Decoder Layer $\times$ Refine Time'' summarizes the refinement configuration.}
\vspace{-3mm}
\centering
\setlength{\tabcolsep}{2.5pt}
\renewcommand{\arraystretch}{1.0}
\resizebox{\linewidth}{!}{
\begin{tabular}{lcc|ccc|ccc|ccc}
\hline
\multirow{2}{*}{Method} &
\multirow{2}{*}{Loss} &
Decoder Layer 
& \multicolumn{3}{c|}{3DPW~\cite{3dpw}}
& \multicolumn{3}{c|}{MOYO~\cite{moyo}}
& \multicolumn{3}{c}{MOYO-HARD}\\
&& $\times$ Refine Time & MPJPE$\downarrow$ & PA$\downarrow$ & PVE$\downarrow$
& MPJPE$\downarrow$ & PA$\downarrow$ & PVE$\downarrow$
& MPJPE$\downarrow$ & PA$\downarrow$ & PVE$\downarrow$\\
\hline
Sparse super. (Ours) & $\mathcal{L}_{refine}$& 6 $\times$ 1
& 61.5 & 38.7 & 73.5
& \bf 85.0 & \bf  51.4 & \bf  91.9
& \bf 90.0 & \bf 61.5 & \bf 102.5\\

Full super. &$\mathcal{L}_{skel}$& 6 $\times$ 1
& \bf 61.3  & \bf 38.6 & \bf 73.3
& 86.4 & 53.2 & 93.9
& 92.1 & 65.2 & 105.6\\

Looped Decoder &$\mathcal{L}_{refine}$& 1 $\times$ 6
& 62.5 & 40.1 & 74.8
& 86.8 & 54.4 & 94.2
& 96.8 & 70.1 & 111.5\\
\hline
\end{tabular}}

\label{tab:pred_compare}
\end{table}

\paragraph{Impact of the encoder's coarse camera translation.}
We compare the encoder's coarse camera translation with a PnP-based camera translation on EMDB~\cite{kaufmann2023emdb}. As shown in Table~\ref{tab:emdb_translation_}, PnP provides only marginal improvement over the encoder prediction while slightly reducing the inference speed from 91.7 FPS to 84.7 FPS. Notably, this comparison favors PnP by using ground-truth 2D keypoint annotations. In practical deployments, where ground-truth keypoints are unavailable, PnP would require an additional 2D keypoint detector, introducing further computational overhead. Therefore, we adopt the encoder's single-forward-pass camera prediction in SKEL-CF.

\begin{table}[t]
    \caption{\textbf{Impact of the encoder's coarse camera translation on EMDB~\cite{kaufmann2023emdb}.} We compare the encoder's coarse camera translation against a PnP-based estimate computed from ground-truth 2D keypoint annotations.}
    \centering
    \resizebox{0.7\columnwidth}{!}{%
        \setlength{\tabcolsep}{3pt}
        \renewcommand{\arraystretch}{0.9}
        \footnotesize
        \begin{tabular}{lccccc}
            \toprule
            Method & L2$\downarrow$ & $t_x\downarrow$ & $t_y\downarrow$ & $t_z\downarrow$ & FPS on H100 \\
            \midrule
            Encoder coarse trans. & 0.4223 & 0.0414 & 0.0294 & 0.4127 & 91.7 \\
            PnP trans. w. G.T. 2D Keypoints & 0.4188 & 0.0403 & 0.0296 & 0.4096 & 84.7 \\
            \bottomrule
        \end{tabular}
    }
    \label{tab:emdb_translation_}
\end{table}

\section{Conclusion}
In this work, we introduced SKEL-CF, a biomechanical skeleton and surface mesh recovery pipeline built upon the SKEL representation. By transforming the high-quality SMPL dataset into a SKEL-based format, termed HMR-SKEL, we established a strong foundation for accurate and generalizable training. Our integration of an explicit camera intrinsic predictor effectively mitigates the limitations of weak-perspective assumptions seen in previous works such as HSMR, enabling robust pose estimation under diverse viewpoints. Furthermore, the proposed coarse-to-fine refinement strategy enables stable and high-fidelity reconstruction results. Overall, SKEL-CF sets a new baseline for biomechanical skeleton and surface recovery, providing a practical and extensible framework for future research in human modeling and motion understanding.

\paragraph{Acknowledgment}
We gratefully acknowledge the financial support from Intellindust, especially Dr. Caizhi Zhu and Xiao Zhou. We also thank the DiDi Infra Team and Great Bay University for providing computational resources. Finally, we sincerely thank the anonymous reviewers for their insightful comments and constructive feedback, which greatly helped improve this paper.

\bibliographystyle{splncs04}
\bibliography{main}

@String(CVPR  = {IEEE Conf. Comput. Vis. Pattern Recog.})

@String(ICCV  = {Int. Conf. Comput. Vis.})

@String(ECCV  = {Eur. Conf. Comput. Vis.})

@String(NeurIPS = {Adv. Neural Inform. Process. Syst.})

@String(ICLR  = {Int. Conf. Learn. Represent.})

@String(TOG   = {ACM Trans. Graph.})

@String(CVPR  = {CVPR})

@String(ICCV  = {ICCV})

@String(ECCV  = {ECCV})

@String(NeurIPS = {NeurIPS})

@String(ICLR  = {ICLR})

@String(TOG   = {ACM TOG})

@inproceedings{xia2025hsmr,
  title={Reconstructing Humans with a Biomechanically Accurate Skeleton},
  author={Xia, Yan and Zhou, Xiaowei and Vouga, Etienne and Huang, Qixing and Pavlakos, Georgios},
  booktitle={CVPR},
  year={2025}
}

@inproceedings{xu2020ghum,
  title={Ghum \& ghuml: Generative 3d human shape and articulated pose models},
  author={Xu, Hongyi and Bazavan, Eduard Gabriel and Zanfir, Andrei and Freeman, William T and Sukthankar, Rahul and Sminchisescu, Cristian},
  booktitle={CVPR},
  year={2020}
}

@inproceedings{
  xu2022vitpose,
  title={Vi{TP}ose: Simple Vision Transformer Baselines for Human Pose Estimation},
  author={Yufei Xu and Jing Zhang and Qiming Zhang and Dacheng Tao},
  booktitle={NeurIPS},
  year={2022},
}

@inproceedings{li2024ohumanoidrobot,
  title={OKAMI: Teaching Humanoid Robots Manipulation Skills through Single Video Imitation},
  author={Li, Jinhan and Zhu, Yifeng and Xie, Yuqi and Jiang, Zhenyu and Seo, Mingyo and Pavlakos, Georgios and Zhu, Yuke},
  booktitle={CoRL},
  year={2024}
}

@article{lan2019albert,
  title={Albert: A lite bert for self-supervised learning of language representations},
  author={Lan, Zhenzhong and Chen, Mingda and Goodman, Sebastian and Gimpel, Kevin and Sharma, Piyush and Soricut, Radu},
  journal={arXiv preprint arXiv:1909.11942},
  year={2019}
}

@inproceedings{tevet2022human,
  title={Human Motion Diffusion Model},
  author={Tevet, Guy and Raab, Sigal and Gordon, Brian and Shafir, Yoni and Cohen-or, Daniel and Bermano, Amit Haim},
  booktitle={ICLR},
  year={2022}
}

@inproceedings{weng2022humannerf,
  title={Humannerf: Free-viewpoint rendering of moving people from monocular video},
  author={Weng, Chung-Yi and Curless, Brian and Srinivasan, Pratul P and Barron, Jonathan T and Kemelmacher-Shlizerman, Ira},
  booktitle={CVPR},
  year={2022}
}

@inproceedings{he2024human2humannoid,
  title={Learning human-to-humanoid real-time whole-body teleoperation},
  author={He, Tairan and Luo, Zhengyi and Xiao, Wenli and Zhang, Chong and Kitani, Kris and Liu, Changliu and Shi, Guanya},
  booktitle={IROS},
  year={2024},
}

@inproceedings{fu2024humanplus,
  title={HumanPlus: Humanoid Shadowing and Imitation from Humans},
  author={Fu, Zipeng and Zhao, Qingqing and Wu, Qi and Wetzstein, Gordon and Finn, Chelsea},
  booktitle={CoRL},
  year=2024
}

@inproceedings{patel2024camerahmr,
  title={Camerahmr: Aligning people with perspective},
  author={Patel, Priyanka and Black, Michael J},
  booktitle={3DV},
  year={2025},
}

@inproceedings{kanazawa2018hmr,
  title={End-to-end recovery of human shape and pose},
  author={Kanazawa, Angjoo and Black, Michael J and Jacobs, David W and Malik, Jitendra},
  booktitle={CVPR},
  year={2018}
}

@inproceedings{smplify,
  title={Keep it SMPL: Automatic estimation of 3D human pose and shape from a single image},
  author={Bogo, Federica and Kanazawa, Angjoo and Lassner, Christoph and Gehler, Peter and Romero, Javier and Black, Michael J},
  booktitle={ECCV},
  year={2016},
}

@inproceedings{kaufmann2023emdb,
  author = {Kaufmann, Manuel and Song, Jie and Guo, Chen and Shen, Kaiyue and Jiang, Tianjian and Tang, Chengcheng and Z{\'a}rate, Juan Jos{\'e} and Hilliges, Otmar},
  title = {{EMDB}: The {E}lectromagnetic {D}atabase of {G}lobal 3{D} {H}uman {P}ose and {S}hape in the {W}ild},
  booktitle = {ICCV},
  year={2023}
}

@article{loper2015smpl,
  title={SMPL},
  author={Loper, Matthew and Mahmood, Naureen and Romero, Javier and Pons-Moll, Gerard and Black, Michael J},
  journal={TOG},
  year={2015},
}

@article{keller2023skin,
  title={From skin to skeleton: Towards biomechanically accurate 3d digital humans},
  author={Keller, Marilyn and Werling, Keenon and Shin, Soyong and Delp, Scott and Pujades, Sergi and Liu, C Karen and Black, Michael J},
  journal={TOG},
  year={2023},
}

@inproceedings{coco,
  title={Microsoft COCO: Common Objects in Context},
  author={Lin, Tsung-Yi and Maire, Michael and Belongie, Serge and Hays, James and Perona, Pietro and Ramanan, Deva and Doll{\'a}r, Piotr and Zitnick, C Lawrence},
  booktitle={ECCV},
  year={2014}
}

@inproceedings{3dpw,
  title={Recovering accurate 3d human pose in the wild using imus and a moving camera},
  author={Von Marcard, Timo and Henschel, Roberto and Black, Michael J and Rosenhahn, Bodo and Pons-Moll, Gerard},
  booktitle={ECCV},
  year={2018}
}

@inproceedings{mahmood2019amass,
  title={AMASS: Archive of motion capture as surface shapes},
  author={Mahmood, Naureen and Ghorbani, Nima and Troje, Nikolaus F and Pons-Moll, Gerard and Black, Michael J},
  booktitle={ICCV},
  year={2019}
}

@article{human36m,
  title={Human3. 6m: Large scale datasets and predictive methods for 3d human sensing in natural environments},
  author={Ionescu, Catalin and Papava, Dragos and Olaru, Vlad and Sminchisescu, Cristian},
  journal={TPAMI},
  year={2013},
}

@inproceedings{detr,
  title={End-to-end object detection with transformers},
  author={Carion, Nicolas and Massa, Francisco and Synnaeve, Gabriel and Usunier, Nicolas and Kirillov, Alexander and Zagoruyko, Sergey},
  booktitle={ECCV},
  year={2020}
}

@inproceedings{lspnet,
  title={Learning effective human pose estimation from inaccurate annotation},
  author={Johnson, Sam and Everingham, Mark},
  booktitle={CVPR},
  year={2011},
}

@inproceedings{kocabas2021pare,
  title={{PARE}: Part attention regressor for 3{D} human body estimation},
  author={Kocabas, Muhammed and Huang, Chun-Hao P and Hilliges, Otmar and Black, Michael J},
  booktitle = {ICCV},
  year={2021}
}

@inproceedings{loshchilov2017decoupled,
  title={Decoupled Weight Decay Regularization},
  author={Loshchilov, Ilya and Hutter, Frank},
  booktitle={ICLR},
  year={2017}
}

@inproceedings{goel20234dhuman,
  title={Humans in 4{D}: Reconstructing and tracking humans with transformers},
  author={Goel, Shubham and Pavlakos, Georgios and Rajasegaran, Jathushan and Kanazawa, Angjoo and Malik, Jitendra},
  booktitle = {ICCV},
  year={2023}
}

@inproceedings{dwivedi2024tokenhmr,
  title={Tokenhmr: Advancing human mesh recovery with a tokenized pose representation},
  author={Dwivedi, Sai Kumar and Sun, Yu and Patel, Priyanka and Feng, Yao and Black, Michael J},
  booktitle={CVPR},
  year={2024}
}

@inproceedings{wang2023refit,
  title={Refit: Recurrent fitting network for 3d human recovery},
  author={Wang, Yufu and Daniilidis, Kostas},
  booktitle={ICCV},
  year={2023}
}

@inproceedings{moyo,
  title={3D human pose estimation via intuitive physics},
  author={Tripathi, Shashank and M{\"u}ller, Lea and Huang, Chun-Hao P and Taheri, Omid and Black, Michael J and Tzionas, Dimitrios},
  booktitle={CVPR},
  year={2023}
}

@inproceedings{smpl-x,
  title={Expressive body capture: 3d hands, face, and body from a single image},
  author={Pavlakos, Georgios and Choutas, Vasileios and Ghorbani, Nima and Bolkart, Timo and Osman, Ahmed AA and Tzionas, Dimitrios and Black, Michael J},
  booktitle={CVPR},
  year={2019}
}

@article{smpl-h,
  title={Embodied hands: modeling and capturing hands and bodies together},
  author={Romero, Javier and Tzionas, Dimitrios and Black, Michael J},
  journal={TOG},
  year={2017}
}

@article{FLAME-HEAD,
  title={Learning a model of facial shape and expression from 4D scans.},
  author={Li, Tianye and Bolkart, Timo and Black, Michael J and Li, Hao and Romero, Javier},
  journal={ACM Trans. Graph.},
  year={2017}
}

@article{SKEL,
  title={From skin to skeleton: Towards biomechanically accurate 3d digital humans},
  author={Keller, Marilyn and Werling, Keenon and Shin, Soyong and Delp, Scott and Pujades, Sergi and Liu, C Karen and Black, Michael J},
  journal={TOG},
  year={2023}
}

@inproceedings{spec-syn,
  title = {{SPEC}: Seeing People in the Wild with an Estimated Camera},
  author = {Kocabas, Muhammed and Huang, Chun-Hao P. and Tesch, Joachim and M{\"u}ller, Lea and Hilliges, Otmar and Black, Michael J.},
  booktitle = {ICCV},
  year = {2021}
}

@article{van2017vqvae,
  title={Neural discrete representation learning},
  author={Van Den Oord, Aaron and Vinyals, Oriol and others},
  journal={NeurIPS},
  year={2017}
}

@inproceedings{li2022cliff,
  title={Cliff: Carrying location information in full frames into human pose and shape estimation},
  author={Li, Zhihao and Liu, Jianzhuang and Zhang, Zhensong and Xu, Songcen and Yan, Youliang},
  booktitle={ECCV},
  year={2022}
}

@inproceedings{kanazawa2019insta,
  title={Learning 3d human dynamics from video},
  author={Kanazawa, Angjoo and Zhang, Jason Y and Felsen, Panna and Malik, Jitendra},
  booktitle={CVPR},
  year={2019}
}

@inproceedings{gu2018ava,
  title={Ava: A video dataset of spatio-temporally localized atomic visual actions},
  author={Gu, Chunhui and Sun, Chen and Ross, David A and Vondrick, Carl and Pantofaru, Caroline and Li, Yeqing and Vijayanarasimhan, Sudheendra and Toderici, George and Ricco, Susanna and Sukthankar, Rahul and others},
  booktitle={CVPR},
  year={2018}
}

@inproceedings{sun2019aic,
  title={Human mesh recovery from monocular images via a skeleton-disentangled representation},
  author={Sun, Yu and Ye, Yun and Liu, Wu and Gao, Wenpeng and Fu, Yili and Mei, Tao},
  booktitle={ICCV},
  year={2019}
}

@inproceedings{pymaf2021,
  title={PyMAF: 3D Human Pose and Shape Regression with Pyramidal Mesh Alignment Feedback Loop},
  author={Zhang, Hongwen and Tian, Yating and Zhou, Xinchi and Ouyang, Wanli and Liu, Yebin and Wang, Limin and Sun, Zhenan},
  booktitle={ICCV},
  year={2021}
}

@article{pymafx2023,
  title={PyMAF-X: Towards Well-aligned Full-body Model Regression from Monocular Images},
  author={Zhang, Hongwen and Tian, Yating and Zhang, Yuxiang and Li, Mengcheng and An, Liang and Sun, Zhenan and Liu, Yebin},
  journal={TPAMI},
  year={2023}
}

@inproceedings{mpi-inf,
  author       = {Dushyant Mehta and
                  Helge Rhodin and
                  Dan Casas and
                  Pascal Fua and
                  Oleksandr Sotnychenko and
                  Weipeng Xu and
                  Christian Theobalt},
  title        = {Monocular 3D Human Pose Estimation in the Wild Using Improved {CNN}
                  Supervision},
  booktitle    = {3DV},
  year         = {2017},

}

@inproceedings{mpii,
  title={2d human pose estimation: New benchmark and state of the art analysis},
  author={Andriluka, Mykhaylo and Pishchulin, Leonid and Gehler, Peter and Schiele, Bernt},
  booktitle={CVPR},
  year={2014}
}

@article{aic,
  title={Ai challenger: A large-scale dataset for going deeper in image understanding},
  author={Wu, Jiahong and Zheng, He and Zhao, Bo and Li, Yixin and Yan, Baoming and Liang, Rui and Wang, Wenjia and Zhou, Shipei and Lin, Guosen and Fu, Yanwei and others},
  journal={arXiv},
  year={2017}
}

@inproceedings{insta,
  title={Learning 3d human dynamics from video},
  author={Kanazawa, Angjoo and Zhang, Jason Y and Felsen, Panna and Malik, Jitendra},
  booktitle={CVPR},
  year={2019}
}

@inproceedings{prohmr,
  title={Probabilistic modeling for human mesh recovery},
  author={Kolotouros, Nikos and Pavlakos, Georgios and Jayaraman, Dinesh and Daniilidis, Kostas},
  booktitle={ICCV},
  year={2021}
}

@inproceedings{shin2024wham,
  title={Wham: Reconstructing world-grounded humans with accurate 3d motion},
  author={Shin, Soyong and Kim, Juyong and Halilaj, Eni and Black, Michael J},
  booktitle={CVPR},
  year={2024}
}

@article{zhu2020deformable,
  title={Deformable detr: Deformable transformers for end-to-end object detection},
  author={Zhu, Xizhou and Su, Weijie and Lu, Lewei and Li, Bin and Wang, Xiaogang and Dai, Jifeng},
  journal={arXiv},
  year={2020}
}

\clearpage

\clearpage
\begingroup

\setcounter{page}{1}

\setlength{\textfloatsep}{6pt plus 1pt minus 2pt}
\setlength{\floatsep}{6pt plus 1pt minus 2pt}
\setlength{\intextsep}{6pt plus 1pt minus 2pt}
\setlength{\abovecaptionskip}{3pt}
\setlength{\belowcaptionskip}{-2pt}
\renewcommand{\topfraction}{0.95}
\renewcommand{\bottomfraction}{0.85}
\renewcommand{\textfraction}{0.05}
\renewcommand{\floatpagefraction}{0.85}
\renewcommand{\dbltopfraction}{0.95}
\renewcommand{\dblfloatpagefraction}{0.85}
\raggedbottom
\makeatletter
\renewcommand\section{\@startsection{section}{1}{\z@}%
                       {-10\p@ \@plus -2\p@ \@minus -2\p@}%
                       {4\p@ \@plus 2\p@ \@minus 1\p@}%
                       {\normalfont\large\bfseries\boldmath
                        \rightskip=\z@ \@plus 8em\pretolerance=10000 }}
\renewcommand\subsection{\@startsection{subsection}{2}{\z@}%
                       {-8\p@ \@plus -2\p@ \@minus -2\p@}%
                       {3\p@ \@plus 2\p@ \@minus 1\p@}%
                       {\normalfont\normalsize\bfseries\boldmath
                        \rightskip=\z@ \@plus 8em\pretolerance=10000 }}
\makeatother

\begin{center}
    {\LARGE \bfseries SKEL-CF: Coarse-to-Fine Biomechanical Skeleton and Surface Mesh Recovery \par}
    \vspace{0.25em}
    {\large Supplementary Material \par}
\end{center}
\vspace{-0.5em}

\setcounter{section}{0}
\setcounter{subsection}{0}
\setcounter{figure}{0}
\setcounter{table}{0}
\setcounter{equation}{0}

\renewcommand{\thesection}{S\arabic{section}}
\renewcommand{\thesubsection}{S\arabic{section}.\arabic{subsection}}
\renewcommand{\thefigure}{S\arabic{figure}}
\renewcommand{\thetable}{S\arabic{table}}
\renewcommand{\theequation}{S\arabic{equation}}

\markboth{Supplementary Material}{Supplementary Material}

\section{Organization}
\label{sec:organization}
Note that additional video demonstrations showcasing both our SKEL-CF visualizations and comparative results with HSMR~\cite{xia2025hsmr} and CameraHMR~\cite{patel2024camerahmr} as well as extended coarse-to-fine visualizations are provided in the project page: \url{https://pokerman8.github.io/SKEL-CF/}.
This appendix document is organized as follows:
\begin{itemize}\setlength{\itemsep}{2pt}\setlength{\topsep}{2pt}\setlength{\parsep}{0pt} %
    
    \item More visual results illustrating our coarse-to-fine refinement strategy are shown in Fig.~\ref{fig:C2F-sup}, as a supplement to Fig.5. 


    
    \item Additional qualitative results comparing our method with the SKEL-based model HSMR~\cite{xia2025hsmr} are presented in Fig.~\ref{fig:compare-hsmr-sup}.
    
    \item Additional qualitative comparisons with the SMPL~\cite{loper2015smpl}-based approach CameraHMR~\cite{patel2024camerahmr} are provided in Fig~\ref{fig:camerahmr-sup}.
    
    \item Details on the MOYO-HARD subset are in Section~\ref{moyo-hard-sup}.


    \item Additional ablation results on 3DPW~\cite{3dpw} and MOYO~\cite{moyo} are reported in Table~\ref{tab:supp_ablation_extended}.

    \item The complete supplementary version of main-paper Table~\ref{tab:smpl_recon_humanmodel} is provided in Table~\ref{tab:supp_smpl_recon_full}.

    \item COCO 2D keypoint results are reported in Table~\ref{tab:2d_pck_coco}.

    \item Section~\ref{sec:attention} analyzes the per-layer attention behavior of Pose, Beta, and Cam tokens.
\end{itemize}

\section{Additional iterative refinement results on LSP-ext~\cite{lspnet}}
\label{iterative-refinement}
Fig.~\ref{fig:C2F-sup} presents additional visual examples from LSP-ext~\cite{lspnet}, highlighting how the refinement process progressively improves pose and shape estimation. 

\begin{figure*}[!htbp]
    \centering
    \includegraphics[width=0.96\textwidth,height=0.66\textheight,keepaspectratio]{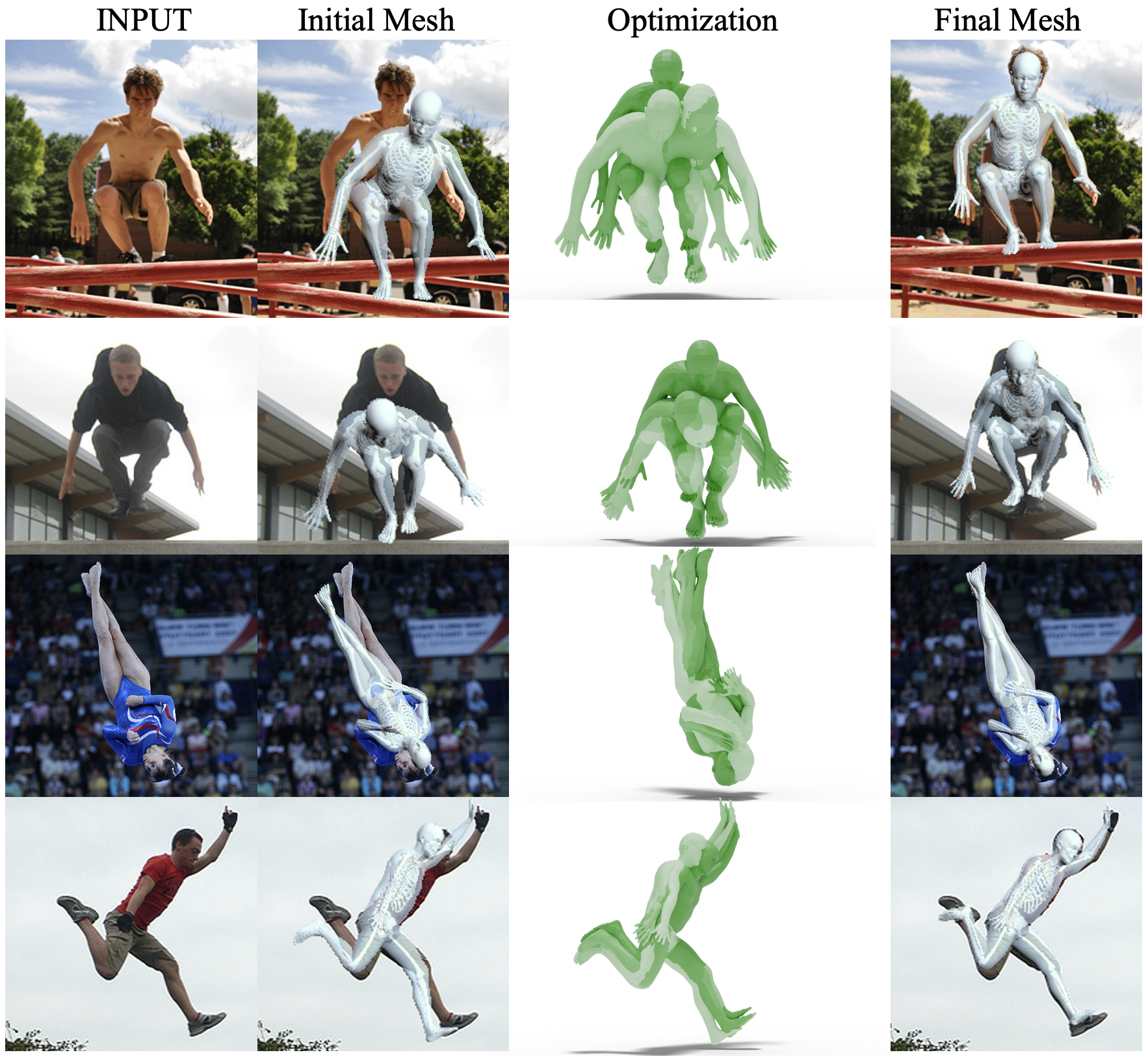}
    \caption{\textbf{Additional iterative refinement results on LSP-ext~\cite{lspnet}.} 
    These examples illustrate how our iterative refinement strategy progressively improves pose and shape estimation, yielding more accurate and stable reconstructions across diverse human poses.}
    \label{fig:C2F-sup}
\end{figure*}

\section{COCO 2D Keypoint Evaluation}
\label{sec:coco-pck-sup}
For completeness, we also report the 2D keypoint results in Table~\ref{tab:2d_pck_coco}. Our approach achieves comparable accuracy under PCK@0.1, indicating that the projected mesh provides a consistent and accurate fit to the 2D image plane. Because the 3D keypoints and the 2D keypoint supervision are not in one-to-one correspondence, improving 3D human mesh quality does not necessarily improve COCO PCK and can sometimes lower it (As shown in Fig.~\ref{fig:rebuttal_coco_2d_}). We therefore treat COCO PCK as a complementary projection metric.

\setlength{\tabcolsep}{8pt} 
\begin{table}[!ht]
\caption{\textbf{2D Keypoint Performance on COCO~\cite{coco}.} 
For completeness, we also report PCK scores of the projected keypoints at thresholds of 0.05 and 0.1, showing that SKEL-CF achieves comparable 2D keypoint quality while maintaining significantly better biomechanical mesh quality.}
\centering
\footnotesize
\resizebox{0.4\linewidth}{!}{
\begin{tabular}{lcc}
\toprule
Method & PCK@0.05 $\uparrow$ & PCK@0.1 $\uparrow$ \\ 
\midrule
PyMAF~\cite{pymaf2021}     & 0.68 & 0.86 \\
CLIFF~\cite{li2022cliff}   & 0.64 & 0.88 \\
PARE~\cite{kocabas2021pare}& 0.72 & 0.91 \\
PyMAF-X~\cite{pymafx2023}  & 0.79 & 0.93 \\
HMR2.0a~\cite{goel20234dhuman} & 0.79 & 0.94 \\
CameraHMR~\cite{patel2024camerahmr} & 0.84 & 0.94 \\
HSMR~\cite{xia2025hsmr} & \textbf{0.85} & \textbf{0.96} \\
SKEL-CF (Ours) & 0.80 & 0.93 \\
\bottomrule
\end{tabular}
}
\label{tab:2d_pck_coco}
\end{table}

\section{\texorpdfstring{Complete SMPL-based Comparison}{Complete SMPL-based Comparison}}
\label{sec:supp-smpl-full}
Table~\ref{tab:supp_smpl_recon_full} provides the complete supplementary version of Table~\ref{tab:smpl_recon_humanmodel} in the main paper, including the MOYO~\cite{moyo} and MOYO-HARD results omitted from the compact main-paper table.

\begin{table}[!ht]
  \caption{\textbf{Complete quantitative comparison of mesh recovery using the SMPL~\cite{loper2015smpl} human model on 3DPW~\cite{3dpw}, EMDB~\cite{kaufmann2023emdb}, SPEC-SYN~\cite{spec-syn}, MOYO~\cite{moyo}, and the more challenging MOYO-HARD dataset.} This table is the complete supplementary version of Table~\ref{tab:smpl_recon_humanmodel} in the main paper, including the MOYO and MOYO-HARD results omitted from the compact main-paper table.}
  \centering
  \begingroup
  \scriptsize
  \setlength{\tabcolsep}{2pt}
  \resizebox{\textwidth}{!}{
  \begin{tabular}{l|ccc|ccc|ccc|ccc|ccc}
    \toprule
    \multirow{2}{*}{\centering \begin{tabular}[c]{@{}c@{}}Method\end{tabular}} & 
    \multicolumn{3}{c}{3DPW~\cite{3dpw}} & 
    \multicolumn{3}{c}{EMDB~\cite{kaufmann2023emdb}} & 
    \multicolumn{3}{c}{SPEC-SYN~\cite{spec-syn}} &
    \multicolumn{3}{c}{MOYO~\cite{moyo}} &
    \multicolumn{3}{c}{MOYO-HARD} \\
    \cmidrule(lr){2-4}
    \cmidrule(lr){5-7}
    \cmidrule(lr){8-10}
    \cmidrule(lr){11-13}
    \cmidrule(lr){14-16}
    & MPJPE$\downarrow$ & PA-MPJPE$\downarrow$ & PVE$\downarrow$ & 
      MPJPE$\downarrow$ & PA-MPJPE$\downarrow$ & PVE$\downarrow$ & 
      MPJPE$\downarrow$ & PA-MPJPE$\downarrow$ & PVE$\downarrow$ &
      MPJPE$\downarrow$ & PA-MPJPE$\downarrow$ & PVE$\downarrow$ &
      MPJPE$\downarrow$ & PA-MPJPE$\downarrow$ & PVE$\downarrow$ \\
    \midrule
    \multicolumn{16}{c}{\textbf{SMPL~\cite{loper2015smpl}-based Approaches}} \\
    \rowcolor{LightGray}
    SPEC~\cite{spec-syn} & 96.5 & 53.2 & 118.5 & 138.9 & 87.7 & 161.3 & 83.5 & 56.9 & 98.9 & - & - & - & - & - & - \\ \rowcolor{LightGray}
    CLIFF~\cite{li2022cliff} & 69.0 & 43.0 & 81.2 & 103.5 & 68.3 & 123.7 & 128.5 & 55.8 & 139.0 & 154.6 & 109.3 & 155.7 & - & - & -\\ \rowcolor{LightGray}
    HMR2.0a~\cite{goel20234dhuman} & 69.8 & 44.4 & 82.2 & 97.8 & 61.5 & 120.0 & 133.3 & 55.8 & 153.0 & - & - & - & - & - & -\\ \rowcolor{LightGray}
    TokenHMR~\cite{dwivedi2024tokenhmr} & 70.5 & 43.8 & 86.0 & 88.1 & 49.8 & 104.2 & 110.5 & 51.8 & 127.6 & - & - & - & - & - & - \\ \rowcolor{LightGray}
    WHAM~\cite{shin2024wham} & 57.8 & 35.9 & 68.7 & 79.7 & 50.4 & 94.4 & - & - & - & - & - & - & - & - & - \\ \rowcolor{LightGray}
    ReFit~\cite{wang2023refit} & 57.6 & 38.2 & 67.6 & 91.7 & 55.5 & 106.2 & 103.6 & 51.3 & 116.3 & - & - & - & - & - & -\\ \rowcolor{LightGray}
    CLIFF~\cite{li2022cliff} & 72.0 & 46.6 & 85.0 & 97.1 & 61.3 & 113.2 & 109.9 & 55.6 & 124.6 & - & - & - & - & - & -\\ \rowcolor{LightGray}
    HMR2.0b~\cite{goel20234dhuman} & 81.3 & 54.3 & 93.1 & 118.5 & 79.2 & 140.6 & 150.7 & 67.6 & 172.9 & 123.3 & 90.4 & 142.2 & - & - & -\\ \rowcolor{LightGray}
    CameraHMR~\cite{patel2024camerahmr} & 62.7 & 38.7 & \textbf{73.4} & 73.2 & \textbf{43.9} & 85.6 & \textbf{66.0} & \textbf{37.0} & \textbf{79.1} & 85.1 & \textbf{50.2} & 95.3 & 96.0 & 61.9 & 111.8 \\ 
    \multicolumn{16}{c}{\textbf{SKEL~\cite{SKEL}-based Approaches}} \\ 
    HSMR~\cite{xia2025hsmr} & 81.5 & 54.8 & - & - & - & - & - & - & - & 104.5 & 79.6 & 120.1 & 120.0 & 97.7 & 140.5 \\ 
    SKEL-CF (Ours) & \textbf{61.5} & \textbf{38.7} & 73.5 & \textbf{72.0} & 44.5 & \textbf{84.7} & 69.4 & 37.1 & 83.4 & \textbf{85.0} & 51.4 & \textbf{91.9} & \textbf{90.0} & \textbf{61.5} & \textbf{102.5} \\
    \bottomrule
  \end{tabular}
  }
  \endgroup
  \label{tab:supp_smpl_recon_full}
\end{table}


\section{Details of MOYO-HARD dataset}
\label{moyo-hard-sup}
The MOYO~\cite{moyo} test set consists of 29 yoga sequences, each recorded simultaneously by eight cameras capturing the same pose from different viewpoints. Each sequence progresses from simple poses to complex ones and then returns to simpler poses, as shown in Fig.~\ref{fig:moyo-sup}. To specifically analyze performance under challenging poses, we focus on the middle portion of each sequence, where the motions are most complex. Accordingly, we construct the MOYO-HARD subset by selecting the middle 50\% of frames from videos captured by Camera~\#1 (front view), resulting in a total of 9,725 images.

\begin{figure*}[!htbp]
    \centering
    \includegraphics[width=0.98\textwidth,height=0.52\textheight,keepaspectratio]{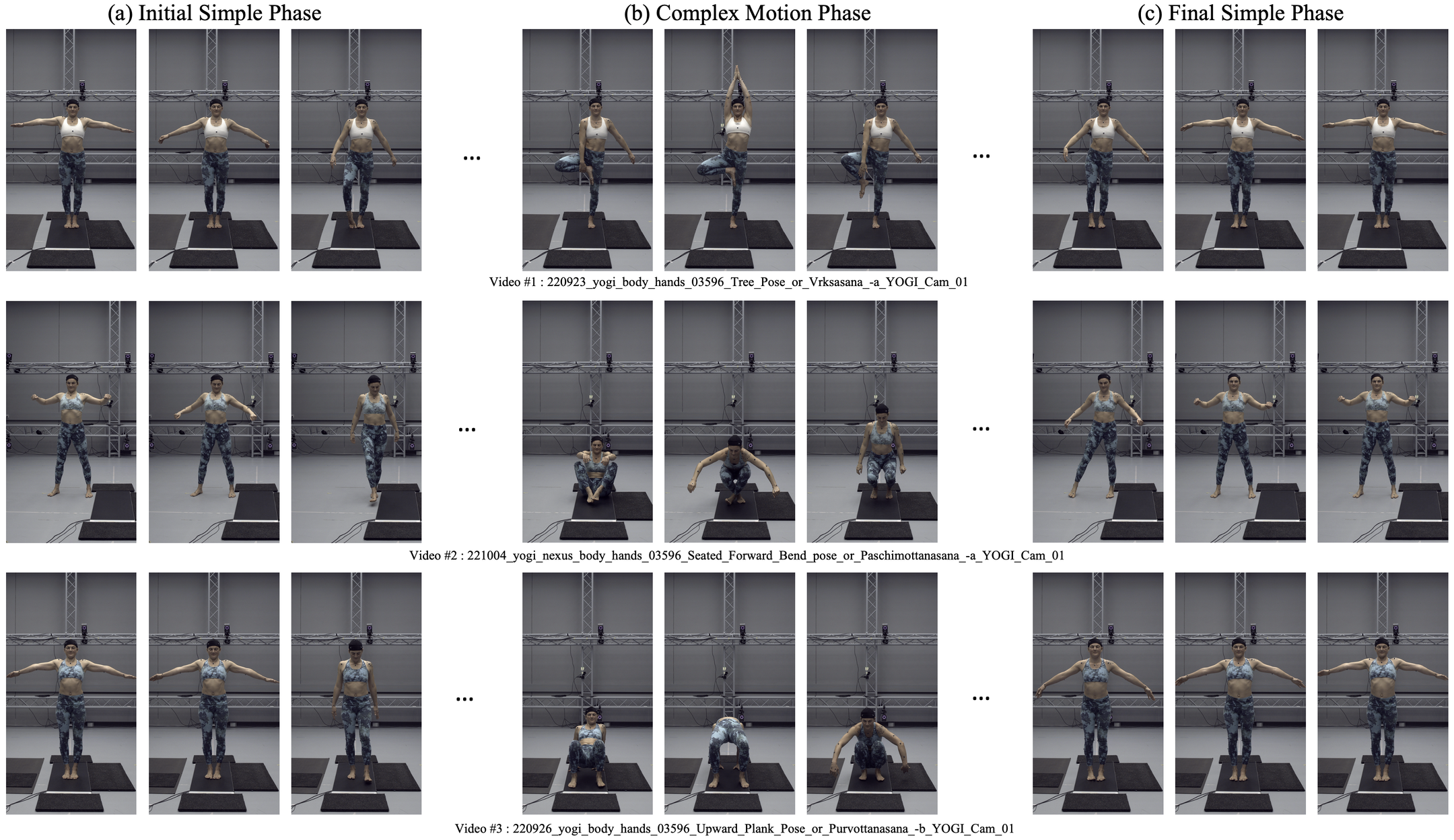}
    \caption{\textbf{Pose Sequence from MOYO.} The video starts and ends with static, low-motion poses, while the middle segment contains diverse, high-complexity movements, reflecting the overall difficulty distribution of the dataset.}
    \label{fig:moyo-sup}
\end{figure*}

\section{\texorpdfstring{Additional Ablation Studies}{Additional Ablation Studies}}
\label{sec:supp-ablation}
Table~\ref{tab:supp_ablation_extended} extends the main-paper ablation to 3DPW~\cite{3dpw} and MOYO~\cite{moyo}, while keeping the MOYO-HARD and COCO~\cite{coco} results for completeness. The trends are consistent with the main analysis: improved HMR-SKEL annotations provide a strong baseline, camera intrinsic modeling improves 3D reconstruction, and the coarse-to-fine and iterative refinement components further improve robustness on challenging poses.

\begin{table}[!htbp]
  \caption{\textbf{Extended ablation study on 3DPW~\cite{3dpw}, MOYO~\cite{moyo}, MOYO-HARD, and COCO~\cite{coco}.}
  \textit{Cam} denotes camera intrinsic modeling,
  \textit{C2F} denotes coarse-to-fine initialization,
  and \textit{Refine} denotes iterative residual refinement.}
  \centering
  \resizebox{\textwidth}{!}{%
  \setlength{\tabcolsep}{1.8pt}
  \renewcommand{\arraystretch}{0.72}
  \tiny
  \begin{threeparttable}
  \begin{tabular}{c|cccc|ccc|ccc|ccc|cc}
    \toprule
    \multirow{2}{*}{Name} & \multicolumn{4}{c|}{Components} & \multicolumn{3}{c|}{3DPW~\cite{3dpw}} & \multicolumn{3}{c|}{MOYO~\cite{moyo}} & \multicolumn{3}{c|}{MOYO-HARD} & \multicolumn{2}{c}{COCO~\cite{coco}} \\
    \cmidrule(lr){2-5}\cmidrule(lr){6-8}\cmidrule(lr){9-11}\cmidrule(lr){12-14}\cmidrule(lr){15-16}
     & Cam & C2F & Refine & Dataset &
     MPJPE$\downarrow$ & PA-MPJPE$\downarrow$ & PVE$\downarrow$ &
     MPJPE$\downarrow$ & PA-MPJPE$\downarrow$ & PVE$\downarrow$ &
     MPJPE$\downarrow$ & PA-MPJPE$\downarrow$ & PVE$\downarrow$ &
     PCK@0.05$\uparrow$ & PCK@0.1$\uparrow$ \\
    \midrule
    \rowcolor{LightGray}
    Baseline (HSMR~\cite{xia2025hsmr}) & \xmark & \xmark & \xmark & HMR2.0~\cite{goel20234dhuman} + SKELify~\cite{xia2025hsmr} & 81.5 & 54.8 & -- & 104.5 & 79.6 & 120.1 & 120.0 & 97.7 & 140.5 & \textbf{0.86} & \textbf{0.96} \\
    Baseline w. our dataset & \xmark & \xmark & \xmark & HMR-SKEL & 65.1 & 39.5 & 77.1 & 88.8 & 53.7 & 97.0 & 103.6 & 67.4 & 121.4 & 0.76 & 0.91 \\
    Ours w.o Cam. & \xmark & \cmark & \cmark & HMR-SKEL & 65.4 & 39.8 & 77.6 & 89.5 & 54.8 & 97.9 & 98.8 & 66.1 & 113.7 & 0.79 & 0.93 \\
    Ours w.o C2F & \cmark & \xmark & \cmark & HMR-SKEL & 61.8 & 38.9 & 73.9 & 84.7 & \textbf{50.6} & \textbf{91.7} & 91.5 & 63.1 & 105.6 & 0.67 & 0.91 \\
    Ours w.o Refine. & \cmark & \cmark & \xmark & HMR-SKEL & 61.6 & \textbf{38.6} & 73.6 & 86.2 & 52.1 & 93.5 & 92.7 & 65.4 & 107.7 & 0.77 & 0.92 \\
    Only Cam & \cmark & \xmark & \xmark & HMR-SKEL & \textbf{61.5} & \textbf{38.6} & \textbf{73.5} & \textbf{84.3} & 51.9 & 92.0 & 92.7 & 66.4 & 107.4 & 0.77 & 0.92 \\
    Ours & \cmark & \cmark & \cmark & HMR-SKEL & \textbf{61.5} & 38.7 & \textbf{73.5} & 85.0 & 51.4 & 91.9 & \textbf{90.0} & \textbf{61.5} & \textbf{102.5} & 0.80 & 0.93 \\
    \bottomrule
  \end{tabular}
  \end{threeparttable}
  }
  \label{tab:supp_ablation_extended}
\end{table}

\section{Visual comparison with SMPLs}
\label{sec:comparsion-smpl-sup}
To more comprehensively evaluate our model, we present in Fig.~\ref{fig:camerahmr-sup} additional visual comparisons extending the results shown in Fig.4 of the main paper. These examples further compare our method with the state-of-the-art SMPL~\cite{loper2015smpl}-based approach, CameraHMR~\cite{patel2024camerahmr}. It indicates that our method can reconstruct more natural pose with accurate anatomical joints.

\begin{figure*}[!htbp]
    \centering
    \includegraphics[width=0.98\textwidth]{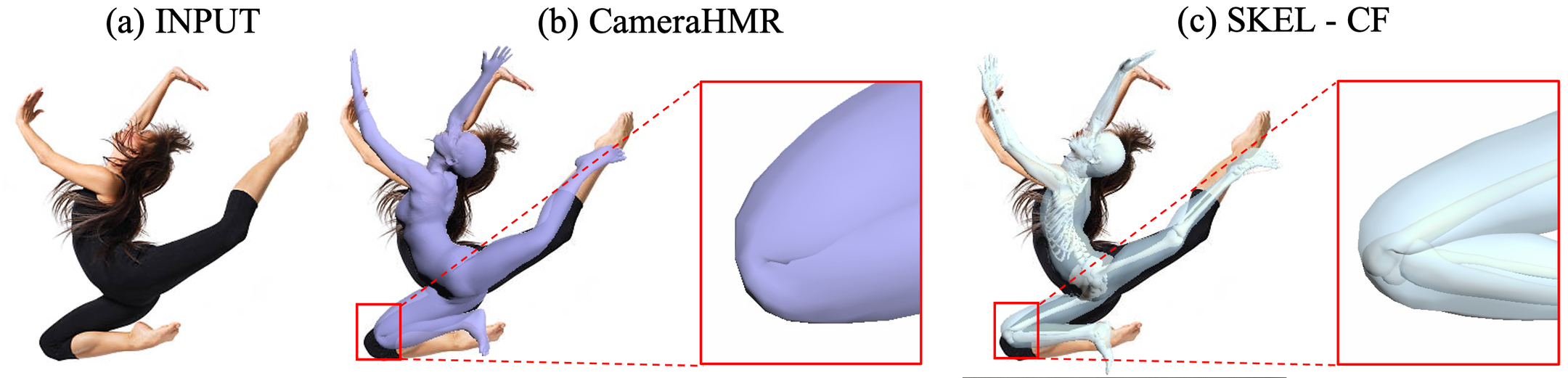}
    \caption{\textbf{Additional comparisons with SMPL~\cite{loper2015smpl}-based models.} 
        Compared with the state-of-the-art SMPL-based method CameraHMR~\cite{patel2024camerahmr}, our approach produces more natural and anatomically consistent joint predictions.}
    \label{fig:camerahmr-sup}
\end{figure*}

\section{Visual comparison with SKELs}
\label{sec:comparsion-skel-sup}

To further compare SKEL~\cite{SKEL}-based approaches, Fig.~\ref{fig:compare-hsmr-sup} shows additional qualitative comparisons with the state-of-the-art HSMR~\cite{xia2025hsmr} pipeline. Unlike HSMR, our method incorporates explicit camera modeling, enabling more reliable and accurate pose reconstruction under varying viewpoints.


\begin{figure*}[!htbp]
    \centering
    \includegraphics[width=0.98\textwidth]{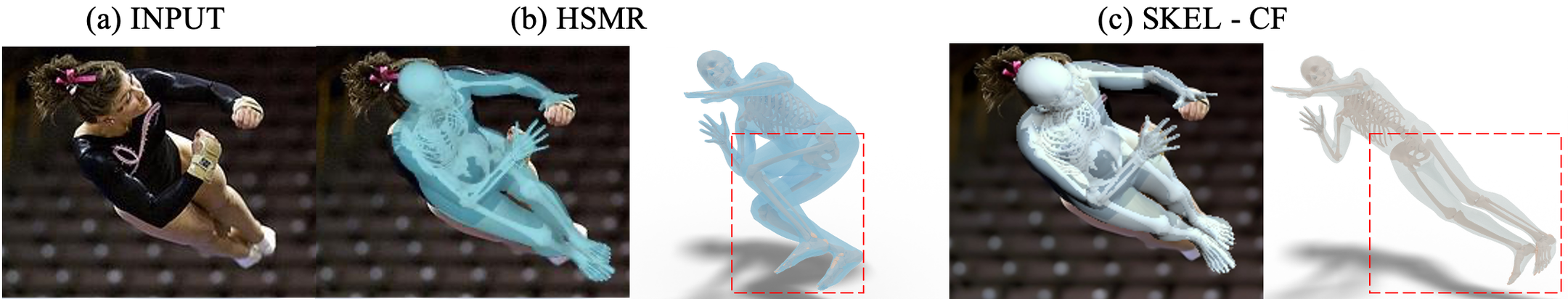}
    \caption{\textbf{Additional comparisons with SKEL~\cite{SKEL}-based models.} 
        Compared with the state-of-the-art HSMR~\cite{xia2025hsmr} method, our approach predicts human poses that are more realistic and accurate.}
    \label{fig:compare-hsmr-sup}
\end{figure*}

\section{Per-Layer Attention Analysis}
\label{sec:attention}
Fig.~\ref{fig:Separate-attention-sup} visualizes the attention regions of the Pose, Beta, and Cam tokens across different decoder layers. In Fig.~\ref{fig:Separate-attention-sup}(a), we show representative examples of the final Pose, Beta, and Cam tokens. We observe that the Pose token primarily attends to the interior body regions, the Beta token focuses on the body silhouette, and the Cam token concentrates on the surrounding environment, as it must aggregate contextual information to estimate the camera extrinsics. In Fig.~\ref{fig:Separate-attention-sup}(b)–(c), we visualize layer-wise attention for the same sample. The Pose token gradually shifts across different body parts, the Beta token evolves from the head region to the full silhouette, and the Cam token expands from the human region to the environment for estimating camera extrinsics.
As shown in Table~\ref{tab:coco_factorization}, our ablation further reveals that layers 1–5 mainly learn 2D translation, while the final layer predominantly captures scale (depth).

\begin{table}[!h]
\centering
\scriptsize
\setlength{\tabcolsep}{4pt} 
\renewcommand{\arraystretch}{1.15}
\caption{\textbf{COCO results under different camera extrinsic factorizations ($s$, $t_x$, $t_y$).} 
(A) Fix final SKEL~\cite{SKEL} parameters and $s$, vary layer-wise $(t_x, t_y)$. 
(B) Fix final SKEL parameters and $(t_x, t_y)$, vary layer-wise $s$.}
\label{tab:coco_factorization}

\begin{tabular}{c|cc|cc}
\toprule
Layer & \multicolumn{2}{c|}{(A) Final Pose + $s$; Layer $(t_x,t_y)$} & \multicolumn{2}{c}{(B) Final Pose + $(t_x,t_y)$; Layer $s$} \\
\midrule
 & PCK@0.05 & PCK@0.1 & PCK@0.05 & PCK@0.1 \\
\midrule
0 & 0.1147 & 0.4156 & 0.0164 & 0.0597 \\
1 & 0.0462 & 0.2493 & 0.0171 & 0.0604 \\
2 & 0.0312 & 0.1876 & 0.0193 & 0.0669 \\
3 & 0.1205 & 0.6088 & 0.0290 & 0.1038 \\
4 & 0.5592 & 0.8837 & 0.0689 & 0.2643 \\
5 & 0.8037 & 0.9361 & 0.8037 & 0.9360 \\
\bottomrule
\end{tabular}

\end{table}

\begin{figure*}[!htbp]
    \centering
    \includegraphics[width=\textwidth]{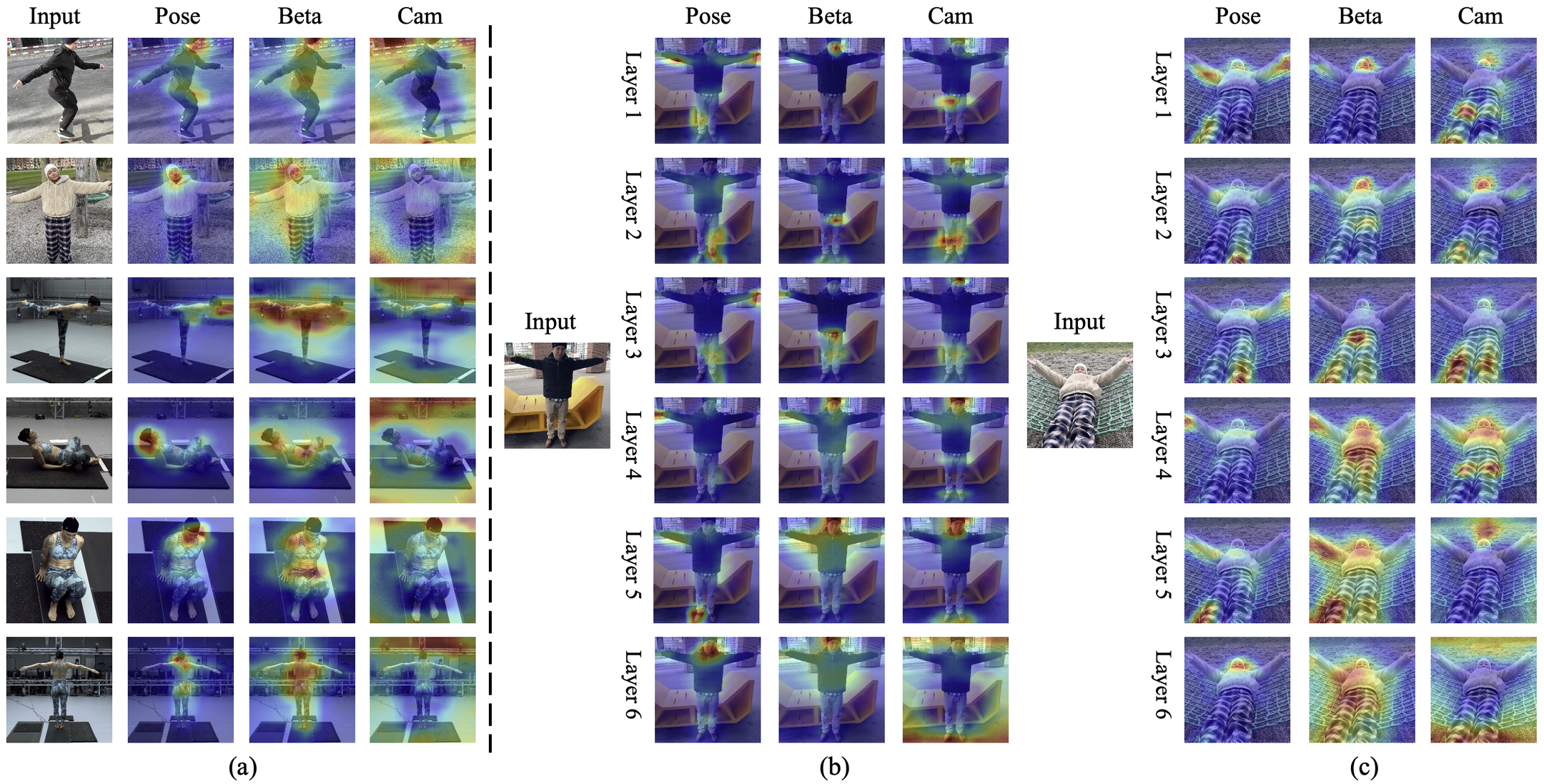}
    \caption{\textbf{Layer-wise attention analysis of Pose, Beta, and Cam tokens.} 
    The Pose token progressively shifts across body parts, the Beta token expands from head to full silhouette, and the Cam token gradually incorporates scene context for estimating camera extrinsics.}
    \label{fig:Separate-attention-sup}
\end{figure*}
\endgroup

\end{document}